# A logical re-conception of neural networks: Hamiltonian bitwise part-whole architecture

**E.F.W.Bowen,[1] R.Granger,[2*] A.Rodriguez[3]**

Dartmouth College

mail@elibowen.net,[1] richard.granger@gmail.com,[2*] antonio.m.rodriguez@dartmouth.edu[3]

**Abstract**

We introduce a simple initial working system in which relations (such as part-whole) are directly represented via an architecture with operating and learning rules fundamentally distinct from standard artificial neural network methods. Arbitrary data are straightforwardly encoded as graphs whose edges correspond to codes from a small fixed primitive set of elemental pairwise relations, such that simple relational encoding is not an add-on, but occurs intrinsically within the most basic components of the system. A novel graph-Hamiltonian operator calculates energies among these encodings, with ground states denoting simultaneous satisfaction of all relation constraints among graph vertices. The method solely uses radically low-precision arithmetic; computational cost is correspondingly low, and scales linearly with the number of edges in the data. The resulting unconventional architecture can process standard ANN examples, but also produces representations that exhibit characteristics of symbolic computation. Specifically, the method identifies simple logical relational structures in these data (part-of; next-to), building hierarchical representations that enable abductive inferential steps generating relational position-based encodings, rather than solely statistical representations. Notably, an equivalent set of ANN operations are derived, identifying a special case of embedded vector encodings that may constitute a useful approach to current work in higher-level semantic representation. The very simple current state of the implemented system invites additional tools and improvements.

## Background and motivation

Artificial neural networks (ANNs; e.g., (Amari, 1967; Grossberg, 1976; McCulloch & Pitts, 1943; Rosenblatt, 1958; Rumelhart & Zipser, 1986; Werbos, 1974; Widrow & Hoff, 1960)) derive from a surprisingly constrained set of specific linear-algebra premises, and present-day networks cleave remarkably closely to these initial formulations.

In ANNs, a cell's response is assumed to be a straightforward function of the sum of products of its inputs and synapses (weights), but the precision of cells and synapses is rarely considered. Yet reducing all high-precision scalars and vectors in these models to primitive binary relations would profoundly change their computations.

We construct a novel form of network that implements a strictly fixed set of primitive binary operations that are interpretable in terms of logic relations. These combine their inputs, weights, and biases to compute a polynomial response that is directly rendered as a Hamiltonian operation on the inputs. The resulting Hamiltonian logic network ("HNet") composes hierarchical representations including complex relations such as part-whole. The resulting learned representations lend themselves to higher-level encodings, which contain relation information of a kind that is usually thought of as more symbolic than statistical.

Standard neural nets and machine learning systems readily represent the "isa" relation (e.g., input X "isa" car), constructing categorization / classification hierarchies composed from "isa" links. It has proven far more challenging to represent other relational types, such as part-of, next-to, above, before (let alone more complex relations such as "capitol of" or "married to").

Much work in AI has been done to try to identify methods for representing and manipulating symbolic aspects of data such as compositional relations: part-whole, contiguity, containment, etc. (e.g., (Hinton, 2021; Holyoak, 2000; Mitchell & Lapata, 2008)). The recent prodigious increase in performance of machine learning systems is predominantly due to the arrival of "transformer" architectures (Vaswani et al., 2017), especially "large language models" (LLMs; e.g., Ramesh, Dhariwal, Nichol, Chu, & Chen, 2022; Ramesh et al., 2021; Zhou et al., 2023) which introduce sequential information supplementing the standard classification-based

* Correspondence to richard.granger@gmail.com

"isa" relation, although much of the information is learned via regression, and remains difficult to inspect or explain (e.g., Mitchell & Krakauer 2023); and the methods are hugely expensive (e.g., Jones 2018).

Much work in AI is focused on hybrid "neuro-symbolic" systems, which attempts to combine statistical information with symbolic, higher-level inference (e.g., Garcez & Lamb, 2020; Günther, Rinaldi, & Marelli, 2019; Holyoak, 2000; Holyoak, Ichien, & Lu, 2022; Kanerva, 2009; Smolensky, McCoy, Fernandez, Goldrick, & Gao, 2022)). Analyses to date suggest shortfalls in current approaches (e.g., (Conwell & Ullman, 2022; Marcus, Davis, & Aaronson, 2022)); the state of the art is rapidly evolving but it is uncertain whether representation of relations and, importantly, the rules of compositionality, can be implemented in systems constructed predominantly as statistical predictive methods (see, e.g., (Thrush et al., 2022)).

The present HNet system i) encodes information in elements that are intrinsically relational rather than just statistical; ii) constructs rich relational hierarchies that may be substantially more readable than current standard systems; and iii) does so by introducing unusual mechanisms that are radically less computationally expensive than in current AI/ML systems; in particular, all operations use only low-precision bitwise arithmetic.

These representations can be submitted to a (supervised) back end that is far simpler (and less costly) than typical extant machine learning and artificial neural network systems. Whereas a simple supervised system (such as an SVM) on its own may achieve a moderate classification rate, we show that HNet representations "boost" the simple supervised system to achieve significantly higher classification rates (see Fig 5 and corresponding text).

Thus substantial costs of typical error-correction mechanisms, and of non-local propagation, may be avoided while retaining the ability to achieve high classification rates. Notably, all HNet calculations are parallel, local, and require only extremely low-precision computation (possibly conforming to several potentially salient computational characteristics of brain circuitry.)

For a standard ANN to capture relations (such as part-whole structure) in data, it typically must step beyond its initial standard statistical calculations, to incorporate increasingly advanced composites (e.g., via convolutions, recurrences, pooling, attention windows); or else the initial steps must be combined with separate symbol-based systems to produce 'hybrids'.

By contrast, the graph-Hamiltonian method presented here uses relations, from the lowest level in the network. Such an approach might mistakenly be perceived to be expensive, if all these low-level operations were individually costly. But the approach in fact is inherently low cost, using binary data and bitwise arithmetic operations throughout its processing steps.

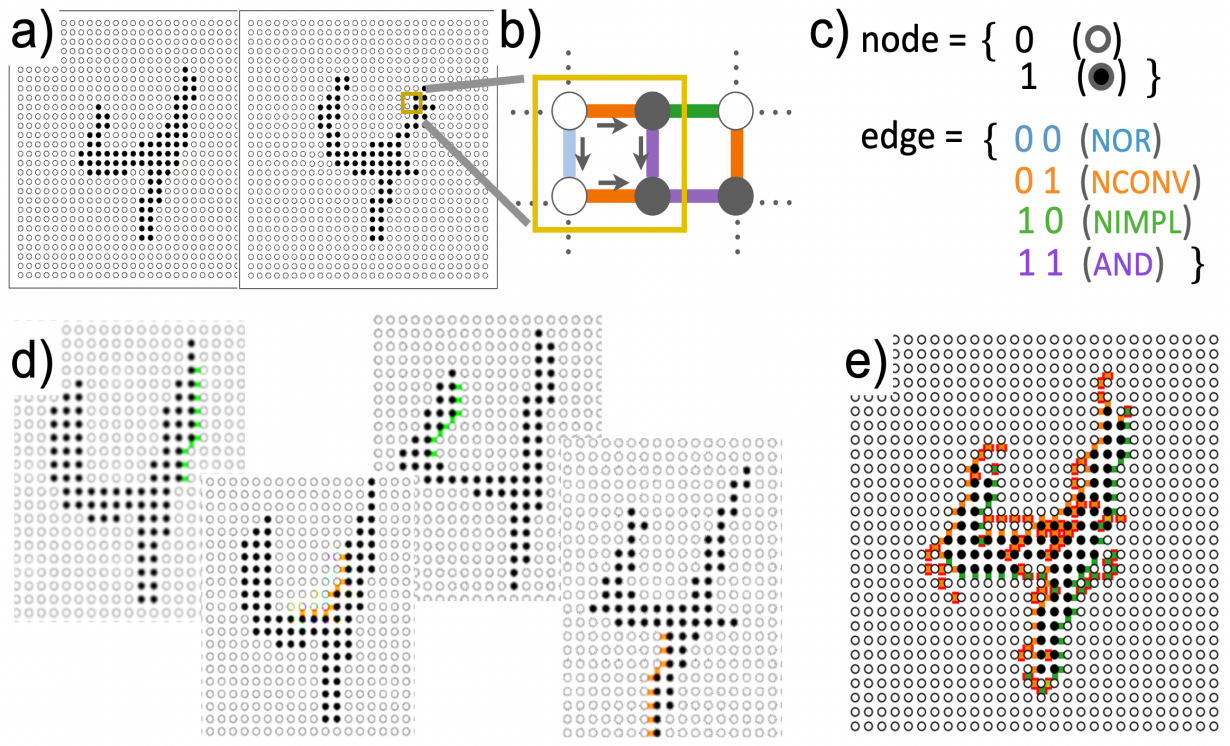

Figure 1. Simple binary relational graph encoding. **a)** Sample MNIST images. **b,c)** Closeup of pixel encoding: binary (black/white) pixels are connected as per directional convention (arrows), by four types of edges (00;01;10;11), corresponding to the logical operations NOR; negative CONVERSE; negative IMPLICATION; AND. **d)** Sample parts in training images. **e)** Sample part matches (colors) and mismatches (red surrounds); matching is via Hamiltonian logic nets (HNets; see text).

## Bitwise relational graph encoding

### Nodes and edges; simple logic notation

We introduce the elements of the HNet method first via examples from the well-studied MNIST dataset (cf. (Fukushima & Miyake, 1982; LeCun et al., 1989; LeCun, Bottou, Bengio, & Haffner, 1998)), and then from a standard credit-card application dataset.

These are emphatically not intended to be examples going head to head against some carefully tailored large system; rather, they are here to introduce fundamental concepts of a new system, building toward more complex examples.

Input data (in this simple pedagogical case, black-and-white MNIST handwritten digits) are encoded in terms of the four possible binary relations among edges (pairs of pixels) in the grid. Each pixel is a binary vertex (node), and each edge is one of the four binary relations (11,10,01,00) (Figure 1; Table 1). (These four binary pairs can be interpreted as True/False logic relations: TT, TF, FT, FF.)

Input data are encoded as simple directed graphs, as in Figure 1. In general, as will be seen, relational binary pairs are not adjacent points in the input space, but in just this initial example (MNIST), we use adjacent pixel pairs for introductory simplicity. In further examples (such as credit applications), arbitrary non-adjacent relations will be used.

## Hamiltonian logic networks (HNets)

Figure 1a shows two sample MNIST digits that have been rasterized and binarized for illustrative simplicity. Figure 1b shows a closeup of a sample region of a digit image, in which adjacent pixels have been coded as binary pairs as in Table 1, either off-off, off-on, on-off, or on-on, as enumerated in Figure 1c. Figure 1d shows encoding of sets of co-occurring binary pairs in this simplified case (connected and adjacent pairs), which will be termed "parts"; each such part can be coded in a proper subspace of the input space, as described in supporting material Appendix A ("Initial part identification").

The general approach is as follows:

- an input graph (or any part thereof) is a "state" $\psi$
- a Hamiltonian $\widehat{H}_\psi$ is produced for each such state;
- an energy value $E_\psi$ can be calculated for any Hamiltonian on any state;
- a state is "recognized" when $E_\psi = 0$

Figure 2 is a simple illustration of the energies of several states.

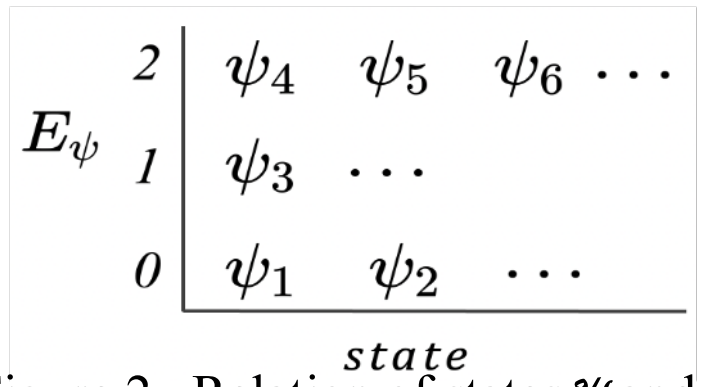

Figure 2. Relation of states $\psi$ and their (Hamiltonian-calculated) energies $E$.

The first two states $\psi_1$, $\psi_2$ in Figure 2 have zero energy values, and thus are successfully "recognized" by the Hamiltonian that produced the figure, whereas all other states have energies of greater than zero, constituting correspondingly poorer matches.

As will be seen, inputs (in this case, MNIST digit images) will be coded in terms of parts (whose construction is described in a later section), and these all are stored as "learned" part representations for training images. When presented with a test image, the test image will also be coded in terms of parts. Each such part has a constructed Hamiltonian, which can be arbitrarily composited into larger HNets. The logical inclusive OR of all stored (learned) parts that match a test image, forms a 'hypothesis' of the test image; these are matched (via energy calculations) against stored train image representations, to identify the best matching training image, and thus the classification of the test image.

| s | | E(s) | | | |
|---|---|---|---|---|---|
| x | y | E(0,0) | E(0,1) | E(1,0) | E(1,1) |
| 0 | 0 | 0 | 1 | 1 | 1 |
| 0 | 1 | 1 | 0 | 1 | 1 |
| 1 | 0 | 1 | 1 | 0 | 1 |
| 1 | 1 | 1 | 1 | 1 | 0 |

Table 2. Energy values E(s) for each of four possible states of binary pairs.

In sum, testing of novel images is performed via three operations:

1] Apply HNets for all learned parts to the test image.
2] Select lowest-energy results.
3] Classify via the OR of those nets' labels ("voting").

All of these operations will first be briefly introduced here, and then elaborated in detail.

For each of the supervised classes in the data (digits 0-9 for this MNIST dataset), training data is encoded in the form of graphs as described, and HNets are produced for each such graph. All operations are performed on just those two data structures: graph encodings, and corresponding Hamiltonians.

A graph encoding of a digit consists of two types of vertices (nodes) and four types of edges. (Additional node types can be envisioned via successive binary subdivisions of such nodes.) In the resulting graph encoding, combinations of edges can be identified as candidate "parts" via the method(s) described in Appendix A.

The HNet for each such part is a matrix whose dimensionality is the square of that of its inputs, and whose values are derived directly from the edges comprising the specific part. As will be seen, these data structures tend to be extremely sparse (i.e., most values are 0).

Moreover, the non-zero values are strictly limited: for the two distinct examples shown here — MNIST digits, and applicants for credit cards — all graph values are binary, and all Hamiltonians can take on only five values: -1, -1/2, 0, 1/2, 1. (The 2x values -2, -1, 0, 1, 2 can be used to yield integer-only arithmetic.)

We begin with the simplest graph: two nodes and a connecting edge. As in Table 1, these take four possible *x,y* values: 00, 01, 10, 11. We define energy scalar $E_r$ (for a given relation state $\underline{r}$) in terms of these *x,y* values of the nodes in the graph plus the associated Hamiltonian $\widehat{H}_r$ , as follows:

$$E_r = [x \;\; y] \;\; \hat{H}_r \begin{bmatrix} x \\ y \end{bmatrix} + k_r$$
$$= [x \;\; y] \;\; \begin{pmatrix} a & b \\ b & c \end{pmatrix} \begin{bmatrix} x \\ y \end{bmatrix} + k_r \qquad \text{(Eq 1)}$$

or in quadratic form:

$$= ax^2 + 2bxy + cy^2 + k_r \qquad (2)$$

(and further simplified for binary values of x and y):

$$= ax + 2bxy + cy + k_r \qquad (3)$$

where x and y are binary data, as described; $a$, $b$, and $c$ are elements of the Hamiltonian (a real Hermitian symmetric matrix such that $\hat{H}_r \in \Re^{2x2}$ and $k_r$ is the energy constant associated with each $\hat{H}_r$ such that $E \geq 0$. For the four binary pairs, the values of $E$ for each state $s$ are defined (Table 2).

Thus, for each of the four binary pairs, the energy can be simply solved for the three variables $a,b,c$ to yield the unique Hamiltonian operator consistent with the specific pair. For instance, for $x,y = 0,0$ (i.e., the relation NOR):

$$E_r(s) = ax + 2bxy + cy + k_r \quad (4)$$

$$\left.\begin{aligned} 0 &= a(0) + 2b(0)(0) + c(0) + k_r \\ 1 &= a(0) + 2b(0)(1) + c(1) + k_r \\ 1 &= a(1) + 2b(1)(0) + c(0) + k_r \\ 1 &= a(1) + 2b(1)(1) + c(1) + k_r \end{aligned}\right\} \begin{aligned} a &= 1 \\ b &= -\frac{1}{2} \\ c &= 1 \\ k_r &= 0 \end{aligned} \quad (5)$$

Thus $a = 1; b = -1/2; c = 1; k_r = 0$ .

Table 3 lists the unit Hamiltonians for each of the four binary pairs. (See also Supplemental Tables 1 and 2). Each HNet is thus formulated to act as a recognizer for the graph from which it is constructed. Applied to a given graph, an HNet yields a scalar energy $E$ that has a value of zero when it recognizes that graph; for any other graphs, the energy will be greater than zero. The category classification $C$ of a given test input can be determined by applying the Hamiltonian for the test input to each of the training inputs, to identify that which produces the lowest energy.

$$C = \text{argmin}_C \left(\vec{x}^T\ \hat{H}_C\ \vec{x}\right) + k_C \quad (6)$$

It is noteworthy that application of $\hat{H}$ conforms with some intuitive and formal logical operations; for instance, negation (NOT) of a given $\hat{H}$ operator is achieved via multiplication by -1. Thus,

$NAND = NOT(AND)$, and

$\hat{H}_{NAND} = -1\,\hat{H}_{AND}$

(Further relations of this kind are discussed in subsequent sections).

$$\hat{H}_{AND} = \begin{bmatrix} 0 & -1/2 \\ -1/2 & 0 \end{bmatrix} \quad \hat{H}_{NCONV} = \begin{bmatrix} 0 & 1/2 \\ 1/2 & -1 \end{bmatrix}$$

$$\hat{H}_{NOR} = \begin{bmatrix} 1 & -1/2 \\ -1/2 & 1 \end{bmatrix} \quad \hat{H}_{NIMPL} = \begin{bmatrix} -1 & 1/2 \\ 1/2 & 0 \end{bmatrix}$$

Table 3: The Hamiltonians of the four binary pairs. (Note that for $\hat{H}_{NOR}$, the $+k$ factor is +0; and for the other three Hamiltonians the $+k$ factor is +1. See Eqs 1-5 and Supplemental Table 2).

## Composing Hamiltonians

Starting from Hamiltonians of simple pairs of two pixels (one edge), we may construct an n-dimensional composite Hamiltonian matrix that corresponds to a graph of arbitrary size; thus HNets can be formulated for any input.

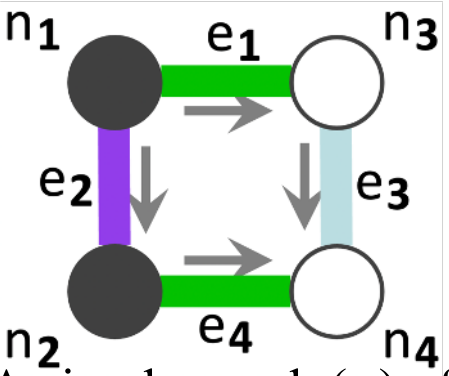

Figure 3. A simple graph ($\alpha$) of four nodes ($n_1$-$n_4$) and edges ($e_1$-$e_4$) (see Fig 1b), to illustrate composing unit Hamiltonians of the four individual pairs into a composite Hamiltonian of this entire four-node graph.

We use a simple four-node graph (Figure 3) to demonstrate how to compose pairwise edges in a graph ($e_1$, $e_2$, $e_3$, and $e_4$ in the figure) into a Hamiltonian for the entire graph, incorporating all its edges. The method immediately extends to enable construction of Hamiltonians for any arbitrarily sized graph (such as MNIST digits).

Given the HNet values for each edge in a graph, a composite HNet is produced, corresponding to the entire graph, by projecting each of the individual unit $\hat{H}$ values (corresponding to single edges) into the 2-dimensional subspace indicated by the variables involved in a given pairwise relation. The resulting $n$-dimensional composite Hamiltonian is obtained by assigning the variables to their appropriate subspaces, and then simply adding the Hamiltonians.

In Figure 3, the Hamiltonian for each edge $e_n$ corresponds to one of the four primary Hamiltonians of Table 3:

$$\begin{aligned} \hat{H}_{e_1} = \hat{H}_{e_4} &= \begin{bmatrix} -1 & 1/2 \\ 1/2 & 0 \end{bmatrix} \quad [k=1] \\ \hat{H}_{e_2} &= \begin{bmatrix} 0 & -1/2 \\ -1/2 & 0 \end{bmatrix} \quad [k=1] \\ \hat{H}_{e_3} &= \begin{bmatrix} 1 & -1/2 \\ -1/2 & 1 \end{bmatrix} \quad [k=0] \end{aligned} \quad (7)$$

Intuitively, to compose these individual edges into the composite graph from Figure 3, each edge is to be situated in its appropriate subset according to the location of its node coordinates:

$$\hat{H}_\alpha = \begin{pmatrix} -1+0 & -1/2 & 1/2 & \\ -1/2 & 0-1 & & 1/2 \\ 1/2 & & 0+1 & -1/2 \\ & 1/2 & -1/2 & 1+0 \end{pmatrix} = \begin{pmatrix} -1 & -1/2 & 1/2 & 0 \\ -1/2 & -1 & 0 & 1/2 \\ 1/2 & 0 & 1 & -1/2 \\ 0 & 1/2 & -1/2 & 1 \end{pmatrix} \quad (8)$$

(where "$\alpha$" denotes the entire four-node graph of Figure 3). (Note edges 1 & 4 both are green).

In general, for an image such as an MNIST digit, the composite HNet will be sparse (corresponding in part to the abundance of blank pixels in a typical digit image).

This resulting Hamiltonian $\hat{H}_\alpha$ can be tested for its "recognition" of the graph $\alpha$ (from Figure 3) by applying the Hamiltonian to the graph:

$$\begin{aligned} E_{\alpha|\alpha} &= \vec{x}_\alpha^T \; \hat{H}_\alpha \; \vec{x}_\alpha + k_\alpha \\ &= (1\,1\,0\,0) \begin{pmatrix} -1 & -1/2 & 1/2 & 0 \\ -1/2 & -1 & 0 & 1/2 \\ 1/2 & 0 & 1 & -1/2 \\ 0 & 1/2 & -1/2 & 1 \end{pmatrix} \begin{pmatrix} 1 \\ 1 \\ 0 \\ 0 \end{pmatrix} + k_\alpha \\ &= 0 \end{aligned} \quad (9)$$

The *0* result indicates that the composite Hamiltonian recognizes or "accepts" the composite graph.

(The same Hamiltonian, tried on a different graph, $\vec{x}_\beta$, would yield a result greater than zero, "rejecting" that graph):

$$\begin{aligned} E_{\beta|\alpha} &= \vec{x}_\beta^T \; \hat{H}_\alpha \; \vec{x}_\beta + k_\alpha \\ &> 0 \end{aligned}$$

## Illustrative applications

### Simple MNIST images example

The network can be applied to standard well-studied tasks of recognition, such as the classic MNIST handwritten-letter dataset containing 28 × 28 pixel (784 pixel) images from ten classes (digits 0-9).

For this simple illustration, the data were preprocessed as follows: i) for the training set, the 32 most prototypical (nearest to the class mean) images per class were selected, s.t. $n = 320$; ii) for the test set, excess images were removed to equalize $n$ = 892 images per class = 8920 total; iii) all images were binarized via pixel intensity thresholding (for these examples, pixel range 0-255; threshold 127).

Each input is rendered in the form of a predefined graph as in Figure 1, such that each node takes the Boolean value of its corresponding element in the input vector, and each edge takes one of four values as a function of its incident nodes, as in Table 1.

A sample training algorithm is shown in Table 4; it is explained as follows (see also Supplemental Tables 4 and 5):

- A graph is defined as described, and each training input is stored. From these two data structures, a composite Hamiltonian is generated. The storage is performed by the function **MEMORIZE(·)** (see Table 4).
  (In the present method, two versions of an input are stored: one with solely all 1,0 (NIMPL) edges (leading edges) and all other edges null (not present); the other with solely all 0,1 (NCONV) edges (trailing edges). The NOR and AND (0,0 and 1,1) edges are typically not stored because they are found to carry little predictive power in typical MNIST images.)
- In **EXTRACT(·)**, each such training image is split into "parts" which are connected components of a new graph capturing the adjacency among edges. For non-image (non topographic) data, the components are defined via statistical regularity rather than connectedness, as will be seen in the "credit card application" illustrative example in the next section.)
- **COMPOSE(·)** transforms a list of edge states into a list of unit Hamiltonians, which then are combined into a single composite Hamiltonian via the composite equations from the previous section. (For topographic data such as images, an additional operation, **CONVOLVE(·)** creates multiple copies of each component, each translated by -2 to +2 in the x and y directions; this simple discrete convolution is performed via a permutation matrix, described in Appendix B).
- Finally, the **ENCODE(·)** operation calculates the energy $\boldsymbol{E}$ of a given $\widehat{H}$ with respect to a give stored representation. When the current input would produce a composite Hamiltonian $\widehat{H}$ identical to a memorized Hamiltonian, or the current input matches the memory, the energy will be zero. Less-good matches will have higher (worse) energies.

```
for i ∈ {1 .. n_train}
   S_train[ :, i]
        ← MEMORIZE(X_train[ :, i], GT1, mask=NIMPL)

for i ∈ {1 .. n_train}
   S_comp[ :, j+1], S_comp[ :, j+2], ... , S_comp[ :, j+m]
        ← EXTRACT(S_train[ :, i], GT1, mode=topog)
   j = j + m

for i ∈ {1 .. n_comp}
   Ĥ [ :, :, i] ← COMPOSE(S_comp[ :, i], GT1)

for j ∈ {1 .. n_train}
   for i ∈ {1 .. n_comp}
      E i,j ← ENCODE( Ĥ [ :, :, j], k[i], X_train[ :, j])
```

Table 4. Pseudocode for an HNet training algorithm as applied to MNIST image data.

Since each stored representation can be paired with a supervised category training label, test inputs can be classified according to the label(s) of the stored representations that are matched with the lowest energies.

Independent of supervised classification, the representations constructed by this HNet algorithm may be of interest in their own right, and they may also be input to various other methods such as other classifiers. For instance, the representations can be provided as input to an arbitrary back end (such as a support vector machine (SVM)) to associate the encoded image with the label.

Figure 4 illustrates sample OR'd part-whole matches of MNIST digits; shown are both matches (colors) and mismatches (colors bordered by red lines) for three digit categories (9, 4, 2).

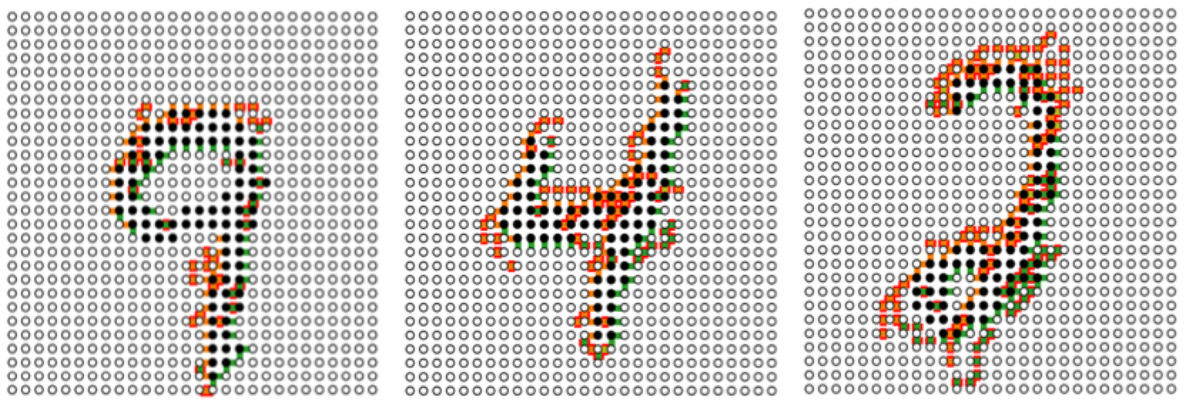

Figure 4. Example OR'd part-whole matches on sample digits 9,4,2. (Color scheme as per Fig 1.)

At testing time, a single vector, *x_test,* is input, corresponding to one (preprocessed) datapoint (image, credit application, etc). The pre-trained model consists of a set of *n_comp* composite Hamiltonians, each of dimensionality mxm where *n_comp* is the number of learned components in the trained model, and *m* is the number of nodes in the graph *G*.

The operation ENCODE measures the energy of *x_test* against each learned component (see Eq 4) using each composite Hamiltonian; these scalar energies are combined into a vector *E* with *n_comp* dimensions.

Finally, the output is passed to the PREDICT(·) function which has been trained (using any standard simple ML method, such as SVMs), to associate energy vectors with class labels.

Alternatively, *E* may be converted from the energy measure to other similarity measures, which then can be used with binary dot product operations. This entails converting each Hamiltonian $\widehat{H}[:, :, i]$ into a "relation vector" $R[:, i]$ whose dimensionality is the number of edges in graph $GT_1$. This is accomplished by first decomposing $\widehat{H}$ into its constituent unit Hamiltonians, each of which is associated with an edge in graph $GT_1$. (The Hamiltonian decomposition is described in Appendix D). Next, the operator associated with the given edge *j* is inferred via Table 3. Finally, we store an integer identifier for this operator in $R[j, i]$.

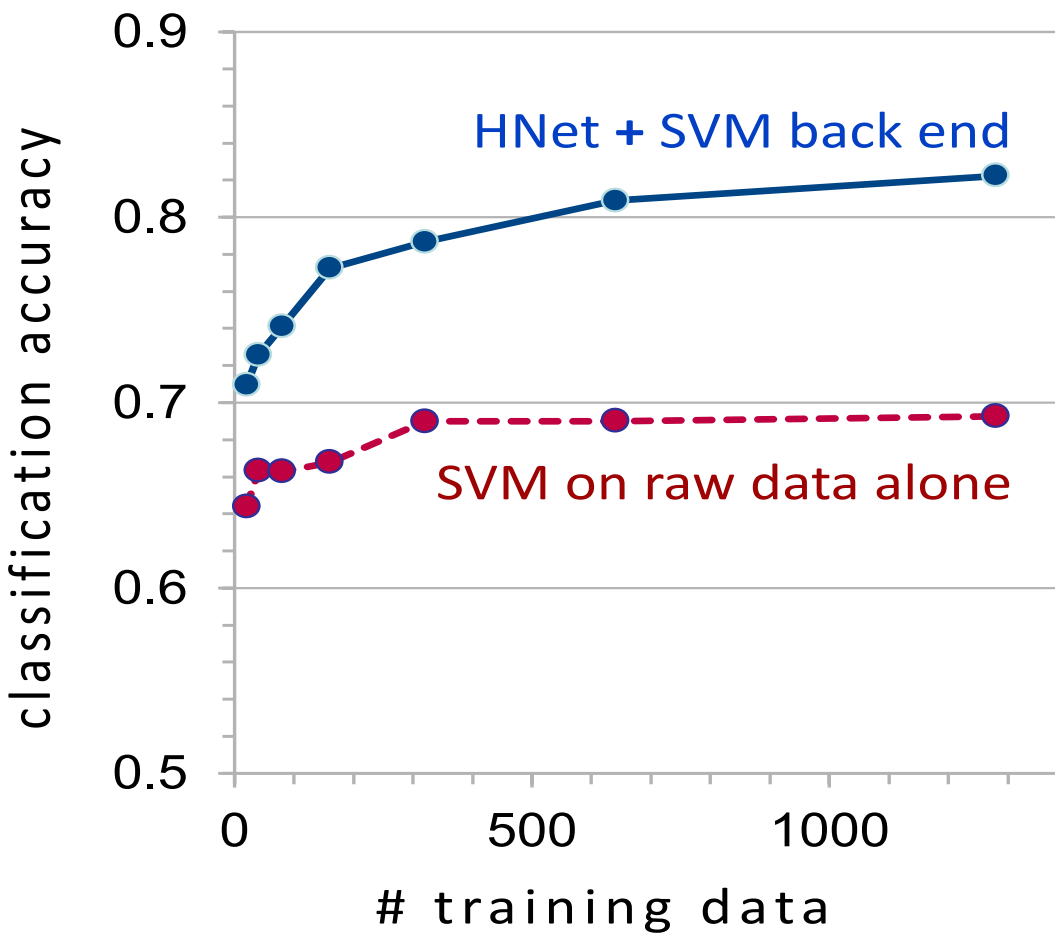

Figure 5. Comparative classification accuracy of a simple SVM back end, with and without HNet processing. SVM achieves 69% accuracy on raw data; HNet with SVM back end achieves 83%, a 14% improvement, due solely to the additional representational richness added by the HNet (see text).

Supervised learning is emphatically not a specific end goal of the work, but rather is compatible with it. The intended primary advantages of HNet are (i) hierarchical relational (such as part-whole) encodings, and (ii) the economy of the bitwise operations and representation encodings, which may enable rich inferences being accomplished with low computational cost. Though it may be tempting to compare classification outcomes against large, expensive, carefully-tuned software, no such attempts are contemplated here. It is useful to note that the performance of straightforward and well-studied engines such as SVMs can be compared when operating on raw data versus on HNet processed data. Figure 5 shows the results of an SVM trained on raw data alone, which achieves a mean classification accuracy rate of 0.69, whereas the same SVM, trained on output produced by the above HNet procedure, has a mean accuracy of 0.83; this roughly 14% immediate improvement apparently arises solely due to the additional representational information constructed by the HNet. It is important to note that this "boosting" effect is achieved at extremely low computational cost.

## Consumer credit application example

Visual image data such as MNIST is organized topographically, i.e., there are neighbor relations among parts of the data, such as the horizontal bar above the slanted vertical line in a numeral "7". Topographic data lends itself to part-whole organization (e.g., the "bar" and "line" in this verbal description of a 7).

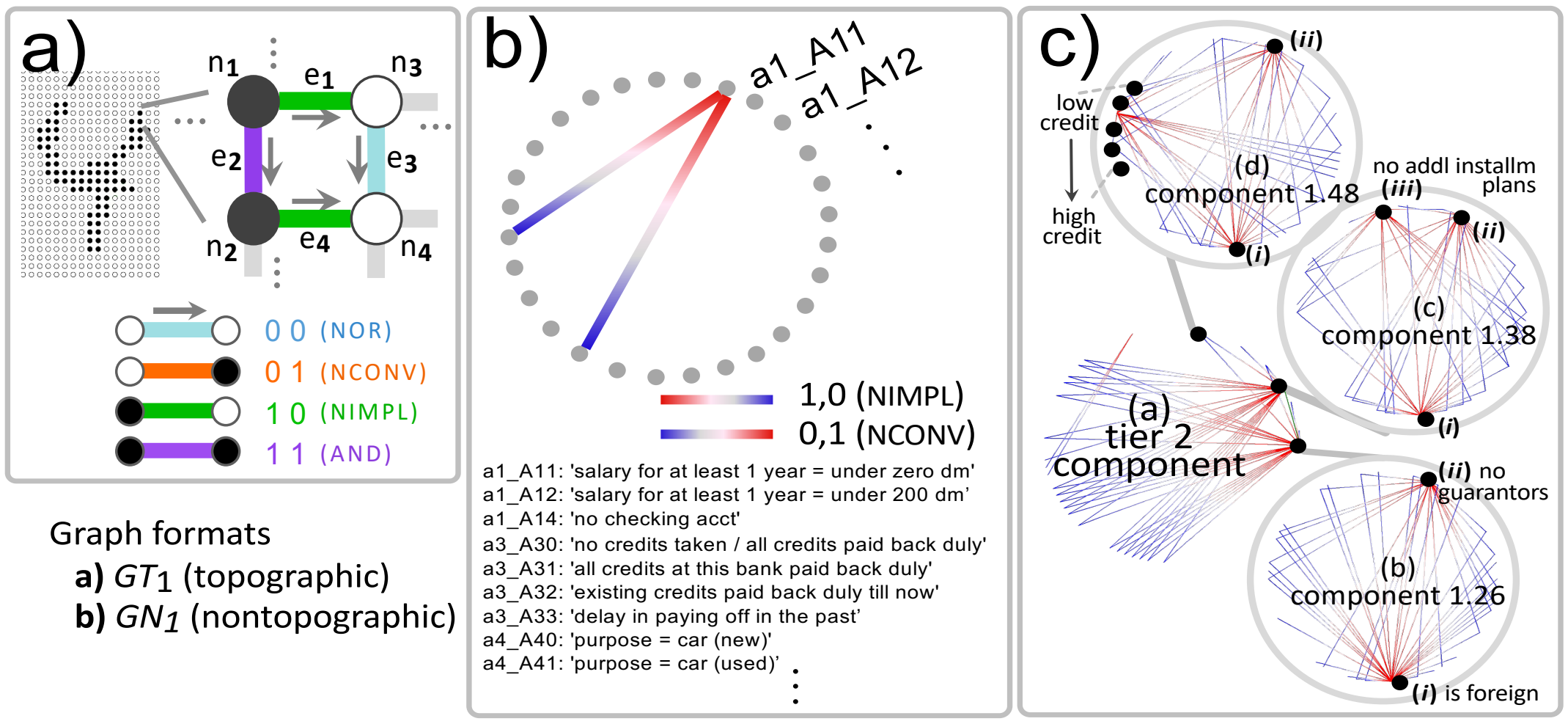

Figure 6. (a,b) Two graph formats used in HNet processing: graph format GT1 (a) for topographic data such as sensory information (e.g., images, sounds); e.g., the MNIST image dataset; graph format GN1 (b) for nontopographic data such as arbitrary abstract attribute-value pairs; examples include consumer credit applications, pricing data, etc. (c) Closeup of credit application data structure (see Supplemental figures 10-12).

Much data occurs in non-topographic form, such as states and capitols; products and prices; etc; these often connote data at a "higher" level of description than pure sensory inputs. These still may have natural hierarchical groupings, such as western or midwestern states; home products vs. automotive vs. industrial products, etc, which may be found to covary in the data, (and are often treated with unsupervised clustering systems).

The HNet formalism readily applies to nontopographic data as well as the topographic examples that already have been discussed; for nontopographic data, inputs are mapped to a graph format termed $GN_1$, distinct from the topographic graph format ($GT_1$) shown in the MNIST examples. Figure 6 illustrates two of the graph formats that the HNet formalism can use (there are additional graph formats for other data structures; these are not discussed in the present paper).

As with topographic examples, the HNet mechanism can construct hierarchies of representations for nontopographic data. These will be illustrated here in the form of "tiers", with input training data initially organized into Tier 1 representations; and those Tier 1 representations further organized into higher-level Tier 2 structures, etc., as will be shown in this section.

Classic examples of nontopographic data abound; in this section we study a straightforward credit card application dataset for simplicity. The emphasis here is not to simply process the discriminative categorization of credit reports, but rather to create rich representations, of primary use for explaining relations identified in the data.

The Statlog German Credit Dataset (Hofmann), available from the UCI Machine Learning Repository (Dua & Graff, 2017), contains two simple classes judging whether a credit application is to be awarded or rejected.

The dataset was preprocessed as follows:

i) Split into training/testing subsets;
ii) Equalized the number of exemplars per class, yielding n[train] = n[test] = 300 (150 per class per set);
iii) All binary variables treated as booleans (true/false);
iv) All categorical variables converted to one-hot codes (further described below);
v) All continuous variables converted into five bins of equal range;

The result is 81 binary features (as listed in Supplemental Table 3).

For this example, the initial input graph $GN_2$ is fully connected, due to the small number of features (input graph nodes) in this credit card dataset. (In general, full connectivity is the natural choice for initial encoding of a nontopographic dataset, for which there is no a priori hypothesis regarding connectivity.)

First, the edge states are stored, i.e., memorized. Unlike illustrative example 1, which contained topographically organized information, we here create a single component for each data point, and store/memorize edges of type 1-1, 0-1, 1-0 (AND, NCONV, NIMPL). Edges of type 0-0 (NOR) are not stored.

An initial candidate Tier-1 model is created by executing independent component analysis (ICA) (Amari, Cichocki, & Yang, 1995) on the set of edge states for the set of training data. Each edge state is represented as a one-hot four dimensional binary vector denoting which of the 4 possible pair-states corresponds to this edge (see Appendix C, and Supplemental Tables 1-3). The resulting edges are stored as the columns of a memory or "state" matrix termed S_t1. ICA is then run on these state vectors; each of the resulting components possesses a weight for each edge, independent in the space of the number of training instances.

Thus dominant sets of edges will be selected for a given component, i.e., edges that co-occur. These are illustrated as thick lines in Figure 7b and c. The weights of the smaller 50% quartile among all components are then set to zero; i.e., the significance of edges is thresholded. Finally, for each edge of each component (among the four possible states), the state with the highest weight "wins" and is stored as the represented weight for that edge in that component.

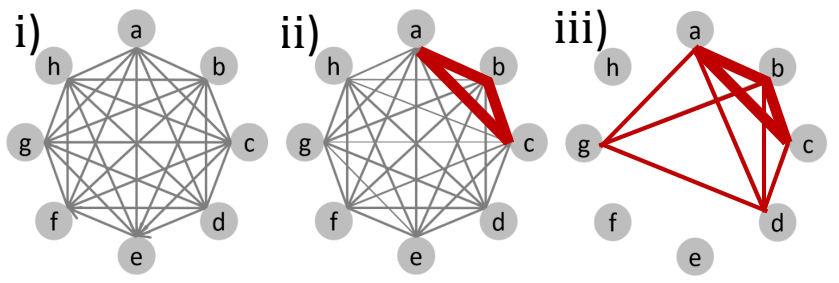

Fig 7. **a)** A fully connected input graph: each node is a credit application variable. **b)** Connections weighted by cross-data-point correlation among variables. Variables a, b, c form a clique of co-varying variables. Edges connecting these variables are likely to be in the same component, and are not removed. **c)** A resulting sample component.

These steps produce a large number of edges per component; these are further sparsified by iteratively removing any given edge that, on average, co-occurs in the training set least often with other of the components' edges. Edges are iteratively removed until no more than 50 edges remain.

Once Tier-1 representations are created, a higher "meta" representational tier is produced such that the nodes in Tier 2 are components of Tier 1, as columns in a matrix *S_t2*. For a given data point, each Tier 2 node's activity is the energy of its corresponding Tier 1 component. In the present example here, Tier 2 components are produced by performing *k*-means clustering on the *S_t2* matrix. As in Tier 1, each Tier 2 edge state (i.e., element in *S_t2* is converted into a one-hot code such that 16 binary numbers represent the state of the edge (see Supplemental Tables 1,2). Each edge is assigned via 1-winner-take-all to one cluster. As before, for each edge of each component amongst the four possible edge states, the most frequent state "wins", and then iterative sparsification is again performed until the number of edges per component is less than 50.

Examples of final Tier 1 and connected Tier 2 representations are shown in Supplemental Figures 10-12, along with brief descriptions of how these may be useful in service of "explanatory" accounts of *what* evaluations may arise from this data as well as *why* given evaluations may arise.

All examples here and all supplemental sections are produced by simple runnable code available at:
https://github.com/DartmouthGrangerLab/hnet

# Identification of equivalences

## Correspondences among systems

A natural question is that of the relation between an HNet's Hamiltonian computations, and those of standard current linear-algebra-based ANN systems. Tables 5 and 6 describe analyses of simple examples both in terms of Hamiltonians and of standard ANN vectors. Shown are the four unit edge types, their corresponding Hamiltonians and the a,b,c values that comprise them (see Table 3), and the corresponding higher-dimensional projection that would be used to operate on these edge types via ANN-like weight matrices. In the math of HNets, for a given edge, a Hamiltonian is stored (via Table 3), and energy may be calculated per equation 1:

$$E = \vec{x}^T \hat{H} \vec{x} + k$$

For that same edge, we may instead create a projection into a four-dimensional space with a specific one-hot vector assigned for each given edge type as in the first two columns of Table 5.

We then can consider unsupervised learning of that four-dimensional one-hot vector. A straightforward method borrows from standard unsupervised "competitive" networks (Rumelhart & Zipser, 1986), by modifying the synapses of a target "winning" dendrite, changing their weights in a direction defined as the difference between the existing synaptic weight, and the value of the input itself (either a 1 or 0 in the four-dimensional "projection" vector from Table 5 here). I.e.,

$$\Delta w_{i,j} = k(x_i - w_{i,j})$$

for a weight on the "winning" target cell in response to input vector $\vec{x}$ . (Note that where the "learning rate" parameter $k$ is 1, then the weight vector $\vec{w}$ becomes equal to the input vector $\vec{x}$ .) To "recognize" this stored memory, a standard weighted sum would be computed:

$$y = \vec{w}^T \ \vec{x} + \text{const}$$

| edge | projection | a b c | $\hat{H}$ |
|---|---|---|---|
| $\begin{matrix}0\\0\end{matrix}$ | $\begin{matrix}1\\0\\0\\0\end{matrix}$ | $\begin{matrix}1\\-1/2\\1\end{matrix}$ | $\begin{pmatrix}1 & -1/2\\ -1/2 & 1\end{pmatrix}$ |
| $\begin{matrix}0\\1\end{matrix}$ | $\begin{matrix}0\\1\\0\\0\end{matrix}$ | $\begin{matrix}0\\1/2\\-1\end{matrix}$ | $\begin{pmatrix}0 & 1/2\\ 1/2 & -1\end{pmatrix}$ |
| $\begin{matrix}1\\0\end{matrix}$ | $\begin{matrix}0\\0\\1\\0\end{matrix}$ | $\begin{matrix}-1\\1/2\\0\end{matrix}$ | $\begin{pmatrix}-1 & 1/2\\ 1/2 & 0\end{pmatrix}$ |
| $\begin{matrix}1\\1\end{matrix}$ | $\begin{matrix}0\\0\\0\\1\end{matrix}$ | $\begin{matrix}0\\-1/2\\0\end{matrix}$ | $\begin{pmatrix}0 & -1/2\\ -1/2 & 0\end{pmatrix}$ |

Table 5: Relationship between Hamiltonians and ANN representations for the four unit edge types. (abc, *H* columns; right): For each edge, the values of a,b,c are defined for the corresponding Hamiltonian. (projection column; middle): For the same edges, the corresponding one-hot vector is shown, effectively projecting two-dimensional edge values into four dimensions, such that each dimension is dedicated to a specific edge type.

| | $\hat{H}$ | composition |
|---|---|---|
| edge 1 (1, 0) | $\begin{pmatrix} -1 & 1/2 \\ 1/2 & 0 \end{pmatrix}$ | $\begin{pmatrix} -1 & 1/2 & \\ 1/2 & 0 & 1/2 \\ & 1/2 & 0 \end{pmatrix}$ |
| edge 2 (1, 1) | $\begin{pmatrix} 0 & 1/2 \\ 1/2 & 0 \end{pmatrix}$ | |

| | projection | composition |
|---|---|---|
| edge 1 (1, 0) | $\begin{pmatrix} 0 \\ 0 \\ 1 \\ 0 \end{pmatrix}$ | $\begin{pmatrix} 0 \\ 0 \\ 1 \\ 0 \\ 0 \\ 0 \\ 0 \\ 1 \end{pmatrix}$ |
| edge 2 (1, 1) | $\begin{pmatrix} 0 \\ 0 \\ 0 \\ 1 \end{pmatrix}$ | |

Table 6. Compositionality in HNet and in a corresponding linear-algebra system (see text). Whereas Hamiltonians for multiple edges are composed as described earlier in this paper, the higher-dimensional one-hot projections of edge vectors are composed by appending them, then projecting them into a higher dimensional space with unusual properties (see text).

(For HNet processing of this same edge, a Hamiltonian would be produced for the edge via Table 3, and it would be recognized via an energy calculation as in Equations 1-3).

Treatment of composite edges entails further projection to higher dimensional spaces. E.g., for two edges, the composing of Hamiltonians would proceed as shown in the "Compositionality" section above, but for the corresponding linear-algebra operation, the two edges would be projected into a correspondingly higher-dimensional space as per Table 6.

In that table, two edges 1-0 and 1-1 are composed first into a composite Hamiltonian as per the compositionality methods described earlier, whereas in the linear-algebra treatment, the two edges would first be projected into four-dimensional space as just described and then the two 4d vectors (for the two edges) would be composed by projection into an eight-dimensional space, consisting of the latter four dimensions appended to the initial four dimensions, such that every four-dimensional subspace is one-hot.

It is worth emphasizing that the higher-dimensional Boolean weight-based version of these computations are formally equivalent to the Hamiltonian version (see Appendix E). Thus, logical relations that can be captured by computations on Hamiltonians can be transferred directly to the weight-based method as projected to higher dimensions. The Hamiltonian formulation thus may have particular value for illuminating formal principles that underlie purely weight-based methods of this specifically constrained kind.

In the weight-based architecture introduced here, inputs are represented by sparse activity in a relatively high dimensional space, with a specific and unusual set of constraints on representational design. Although this design is not typical of standard backprop or deep-backprop architectures, the design nonetheless can be seen to have points of correspondence with a rich literature in which high-dimensional vectors are intended to encode symbolic content, such that the resulting "symbols" may nonetheless then be operated on by numeric rules of the kinds used in ANNs. Key examples are embeddings, as initially in Word2vec, and then in BERT, GloVe, GPT-3, DALL-E, and several additional transformer models, which are of intensive current interest in ML and ANNs (Brown et al., 2020; Mikolov, Sutskever, Chen, Corrado, & Dean, 2013; Pennington, Socher, & Manning, 2014). In general, possible links of this kind, between symbolic and sub-symbolic representations, are the subject of very active ongoing research (see, e.g., (Garcez & Lamb, 2020; Günther, Rinaldi, & Marelli, 2019; Holyoak, 2000; Holyoak, Ichien, & Lu, 2022; Kanerva, 2009; Smolensky, McCoy, Fernandez, Goldrick, & Gao, 2022)) and bear some resemblances with other, quite different, logic-based computational designs (e.g., (Parsa et al., 2022)).

## From hierarchical statistics to abduced symbols

It is perhaps useful to envision some of the ongoing developments that are arising from enlarging and elaborating the Hamiltonian logic net architecture. As yet, no large-scale training whatsoever has gone into the present minimal HNet model; thus far it is solely implemented at a small, introductory scale, as an experimental new approach to representations. It is conjectured that with large-scale training, hierarchical constructs would be accreted as in large deep network systems, with the key difference that, in HNets, such constructs would have relational properties beyond the "isa" (category) relation, as discussed earlier.

Such relational representations lend themselves to abductive steps (McDermott 1987) (or "retroductive" (Pierce

1883)); i.e., inferential generalization steps that go beyond warranted statistical information. If John kissed Mary, Bill kissed Mary, and Hal kissed Mary, etc., then a novel category ₵X can be abduced such that ₵X kissed Mary.

Importantly, the new entity ₵X is not a category based on the features of the members of the category, let alone the similarity of such features. I.e., it is not a statistical cluster in any usual sense. Rather, it is a "position-based category," signifying entities that stand in a fixed relation with other entities. John, Bill, Hal may not resemble each other in any way, other than being entities that all kissed Mary. Position-based categories (PBCs) thus fundamentally differ from "isa" categories, which can be similarity-based (in unsupervised systems) or outcome-based (in supervised systems). PBCs share some characteristics with "embeddings" in transformer architectures.

Abducing a category of this kind often entails overgeneralization, and subsequent learning may require learned exceptions to the overgeneralization. (Verb past tenses typically are formed by appending "-ed", and a language learner may initially overgeneralize to "runned" and "gived," necessitating subsequent exception learning of "ran" and "gave".)

Simple abductive steps are generated as the HNet constructs hierarchies:

| Successive epochs |
|---|
| i) Cluster together multiple similar Tier ***n+1*** instances<br>e.g., (LEFT x1 y1), (LEFT x2 y2), (LEFT x3,y3), … |
| ii) Create Tier ***n+2*** element that ORs the arguments of the Tier ***n+1*** instances.<br>e.g. (LEFT (OR x1 x2 x3) (OR y1 y2 y3)) |
| iii) Abduce Tier ***n+3*** ("type") arguments (₵,$) that overgeneralize from the Tier ***n+2*** OR'd ("token") arguments.<br>e.g., ($LEFT ₵X ₵Y) |

Initial versions of these abductive methods were implemented and applied to the CLEVR image dataset (Sampat et al., 2021). Figure 8 illustrates simple instances of the resulting abduced entities, both objects (categories; ₵A) and sequential relations ($M), and compositions of these.

We conjecture that ongoing hierarchical construction of such entities can enable increasingly "symbol-like" representations, arising from lower-level "statistic-like" representations. Figure 9 illustrates construction of simple "face" configuration representations, from exemplars constructed within the CLEVR system consisting of very simple eyes, nose, mouth features. Categories (₵) and sequential relations ($) exhibit full compositionality into sequential relations of categories of sequential relations, etc.; these define formal grammars (Rodriguez & Granger 2016; Granger 2020). Exemplars (a,b) and near misses (c,d) are presented, initially yielding just instances, which are then greatly reduced via abductive steps (see Supplemental Figure 13).

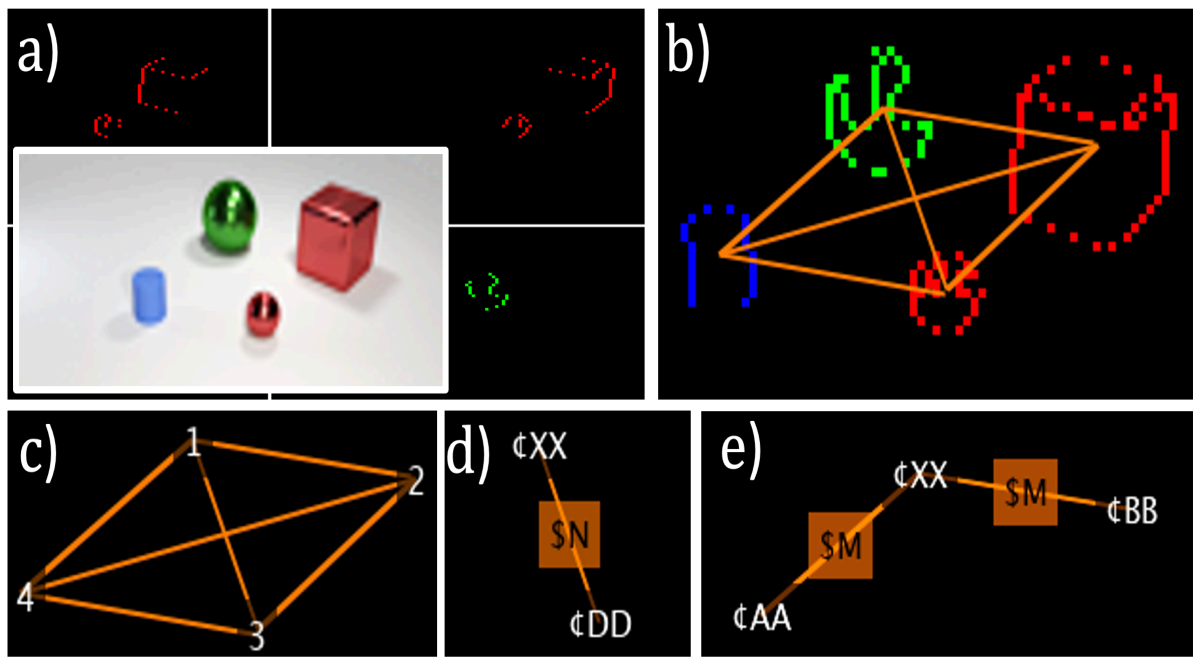

Figure 8. Image features (a; inset) are rendered into individual entities (b) with initial individual relations (c). After repeatedly occurring as a statistical class, they are abduced (d,e) to (possibly overgeneralized) position-based categories of objects (₵XX, ₵DD) and sequential relations ($M, $N). (See text).

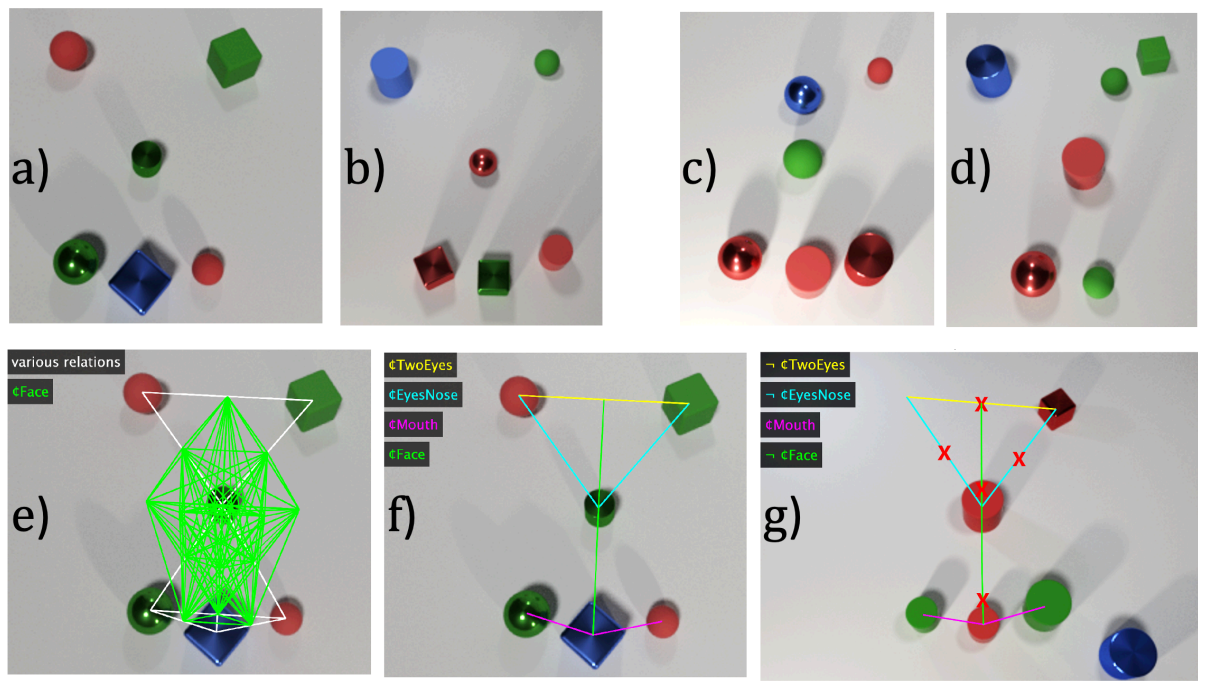

Figure 9. Relations in instances and non-instances of "faces" in CLEVR. As examples (a,b) and near-misses (c,d) are presented, a statistical rendering of the class of faces comes to contain many individual links (e) which are greatly reduced via simple abductive steps to match (f) or fail to match (g) new instances (see text).

## Discussion

A system has been described that operates on arbitrary data, allowing both unsupervised and supervised capabilities, differing from standard extant systems by inherently identifying part-whole relations in data, and constructing logical (and hence potentially more interpretable) representations, all via radically low-precision, intrinsically massively parallel computational operations, in an energy-based architecture. This Hamiltonian bitwise logic network, or HNet, identifies lowest-energy "matching" states, combining statistical information together with instance-based recognition of logical edges in a directed graph formalism.

The algorithm lends itself to direct implementation in appropriate parallel hardware. The values of nodes at each time step can be calculated independently. The algorithm's elapsed processing time depends on the number of edges in the data, i.e., O(n_edges = trace of degree matrix). With a single processor, processing time is O(average n_edges), growing linearly with the number of edges. If each node is assigned to a separate parallel processing element, average computation is O(n_edges/n_nodes = avg. degree of nodes). It is notable that these unusual scalability characteristics differ drastically from standard methods and from other graph methods (e.g., probabilistic graphs and graph NNs), which in general do not resemble HNets; some potentially interesting cases entail the use of probabilistic graphical models (PGMs) that have been adapted to some potential symbolic cases (Koller & Friedman 2009; Rosenbloom 2011).

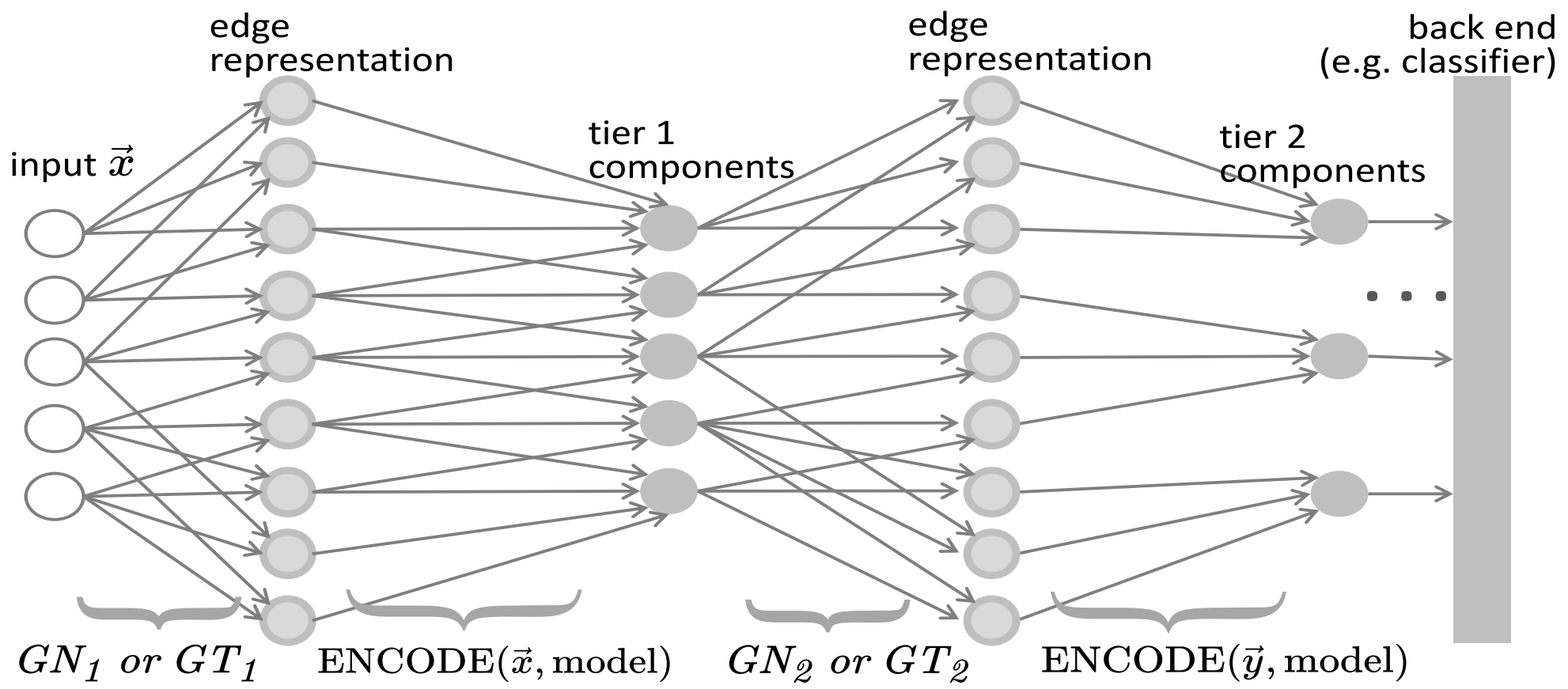

Figure 10. HNet architecture, consisting of successive "layers" that do not correspond at all to typical ANN or deep network layers. Inputs, either topographic (e.g., sensory input, such as MNIST) or nontopographic (abstract value-attribute vectors, such as the credit-card application dataset), are rendered as graphs (GT or GN, respectively), as described in the text. The resulting edge representations are ENCODEd as in the text, into "tier 1" components. All such components can then be rendered as new graphs, whose edges can again be ENCODEd as components, now tier 2 in the growing hierarchy (which can continue with additional tiers, as indicated by ellipsis at right). The resulting representations can be input to any back end mechanism such as a classifier (SVM, *k*-nearest neighbor, simple neural network, etc.). The enriched representation can enable higher classification scores than a classifier alone would attain (see Figure 5), and enable computationally far less costly classifiers, as well as identifying potential symbol-like relations (see text).

Longstanding customary elements of most typical artificial neural networks are entirely discarded here: there is no use of gradient descent / hill climbing; no long-distance propagation of information; no assumption of simple synaptic summation. By contrast, the method incorporates some characteristics that are highly unusual in machine learning or ANNs, but are highly valuable for scaling, especially the use of solely low-precision calculations.

Of particular note are the simple relational elements of the graph representations, including possible abstract symbolic structures. Whereas typical ANNs tend largely to impose additional structure (e.g., convolutions, pooling, etc.) as add-ons to simpler statistically-based components, the graph edges of the HNet formalism contain structural information all the way down at the lowest level of representation.

The two initial pedagogical examples presented here are standard well-studied machine learning and ANN tasks: MNIST digit character recognition, and application for a consumer credit card. The former consists of topographic data, as does any sensory input (images, sounds, etc.), whereas the latter is nontopographic, as is typical for most non-perceptual datasets containing arbitrary attribute-value pairs. In both cases, processing these data can lead to generation of categories beyond typical "isa" links, to generalized position-based categories; as mentioned, these generalize to families of formal grammars (Granger 2020); ongoing studies are pursuing this generalization.

The HNet formalism is not presented as an intended substitute for existing ANN and ML systems; it is a novel method for producing representations, by identifying structure not just among data, but *within* data (e.g., doors are part of cars; handles are part of doors, etc.). The method thus builds richer representations than those typically present in current systems.

The aim, throughout the work, is not to compare performance against any particular machine learning or neural net system (whose performance of course can be substantially modified by many extraneous add-on methods; see, e.g., (Serre, 2019)). Rather, the goal is to illustrate key atypical characteristics of this novel algorithm, especially:

- graph encoding via four specific types of edges;
- “part” identification by matched edge types;
- Hamiltonian HNets for edge recognition;
- full compositionality of HNets;
- solely low-precision arithmetic throughout;
- all local operations; no propagation;
- interpretable, ‘explainable’ representations.

Many of these features are unusual in typical ML/ANN systems. Taken together, they are a thoroughgoing departure from standard approaches, raising the possibility of alternate stratagems for highly scalable, hardware-ready data processing algorithms.

The early system presented here has many characteristics that can be substantially improved upon. For instance, all examples still are of small size on modest datasets; no large-scale constructs have been built or tested. Moreover, the system currently relies on multiple possible component-identification methods (such as ICA); no methodical examination has yet been carried out for optimization of such methods across various applications domains.

In addition, the hierarchical characteristics of the system, although promising, are as yet largely underexplored: it is not known what large-scale structures may be generated were the system to be applied to highly structured data in which concepts (e.g., shape, color, creditworthiness, socio-economic class) might be identified, and it is unknown as yet what capabilities may emerge that enable more “symbolic” or rule-based inferences, beyond those initially implemented here.

Even at this early stage, the system presents a demonstrated method for encoding relations in a way that is currently elusive and highly expensive in extant systems, and it does so straightforwardly and radically inexpensively, via tools that are unusually well suited to hardware implementation.

It is hoped that the simple intrinsic relational elements of HNet graph data structures, and the formulation of Hamiltonians that treat these graphs as computable elements, as well as the markedly low computational processing costs, may offer a possible starting point for further approaches to the representation of relations beyond “isa,” incorporating arbitrary relations that enable a direct crossover from purely low-level (e.g., statistical) systems to symbolic computation.

## Acknowledgments

The work reported herein was supported in part by funding from the Office of Naval Research.

## References

Amari, S. (1967). A theory of adaptive pattern classifiers. *IEEE Trans. Electronic Computers 3*, 299-307.

Amari, S., Cichocki, A., & Yang, H. (1995). *A new learning algorithm for blind signal separation*. Advances in Neur Info Proc Sys

Brown, T., Mann, B., Ryder, N., Subbiah, M., Kaplan, J., Dhariwal, P., …. (2020). *Language models are few-shot learners.* NeurIPs 2020. arxiv.org/abs/2005.14165

Conwell, C., & Ullman, T. (2022). *Testing relational understanding in text-guided image generation*. arXiv:2208.00005

Dua, D., & Graff, C. (2017). UCI machine learning repository. Retrieved from archive.ics.uci.edu/ml archive.ics.uci.edu/ml

Fukushima, K., & Miyake, S. (1982). Neocognitron: A self-organizing neural network model for a mechanism of visual pattern recognition. . In *Competition and coöperation in neural nets.* (pp. 267-285). Berlin: Springer.

Garcez, A., & Lamb, L. (2020). *Neurosymbolic AI: The 3rd wave*. arXiv:2012.05876v2

Granger R (2020). Toward the quantification of cognition. arXiv:2008.05580

Grossberg, S. (1976). Adaptive pattern classification and universal recoding I *Biological Cybernetics, 23*, 121-134.

Günther, F., Rinaldi, L., & Marelli, M. (2019). Vector-space models of semantic representation from a cognitive perspective. *Perspectives on Psychol Sci, 14*, 1006-1033. doi: doi.org/10.1177/1745691619861372

Hinton, G. (2021). *How to represent part-whole hierarchies in a neural network.* arxiv:2102.12627

Holyoak, K. (2000). The proper treatment of symbols. In *Cognitive dynamics*, MIT Press (p.229)

Holyoak K, Ichien N, Lu H. (2022). From semantic vectors to analogical mapping. *Psych Sci*. doi:10.1177/09637214221098054

Jones N. (2018) How to stop data centres from gobbling up the world’s electricity. *Nature 561: 163-166,* https://doi.org/10.1038/d41586-018-06610-y

Kanerva, P. (2009). Hyperdimensional computing. *Cogn Comput, 1*, 139-159. doi:10.1007/s12559-009-9009-8

Koller D, Friedman N (2009). Probabilistic graphical models. Cambridge MA: MIT Press

LeCun, Y., Boser, B., Denker, J., Henderson, D., Howard, R., Hubbard, W., & Jackel, L. (1989). *Handwritten digit recognition with a back-propagation network.* Paper presented at the Adv. Neural Information Proc Sys 2

LeCun Y, Bottou L, Bengio Y, Haffner P (1998) Gradient-based learning applied to document recognition. *IEEE 86*:2278-2342.

Marcus, G., Davis, E., & Aaronson, S. (2022). *A very preliminary analysis of DALL-E 2*. arXiv:2204.13807.

McCulloch, W., & Pitts, W. (1943). A logical calculus of the ideas immanent in nervous activity. *Bull. Math Biol. 5*: 115-133.

McDermott D (1987) A critique of pure reason. Comp. Intell 3:151-160

Mikolov, T., Sutskever, I., Chen, K., Corrado, G., & Dean, J. (2013). *Distributed representations of words and phrases and their compositionality.* Adv. Neural Information Proc Sys (NIPS 2013)

Mitchell, J., & Lapata, M. (2008). *Vector-based models of semantic composition*. Paper presented at the ACL-08: HLT.

Mitchell M, Krakauer D (2023) The debate over understanding in AI's large language models. arXiv:2210.13966v3

Molnar, C. (2022). *A guide for making black box models explainable*. https://christophm.github.io/interpretable-ml-book/

Parsa, A., Wang, D., O'Hern, C., Shattuck, M., Kramer-Bottiglio, R., & Bongard, J. (2022). *Evolving programmable computational metamaterials* Paper presented at the GECCO '22

Peirce C (1883) A theory of probable inference. In "Studies in Logic" Little, Brown (Boston) pp.126-181.

Pennington, J., Socher, R., & Manning, C. (2014). *GloVe: Global vectors for word representation.* Conf Empirical Methods in NLP.

Ramesh, A., Dhariwal, P., Nichol, A., Chu, C., & Chen, M. (2022). *Hierarchical text-conditional image generation with clip latents*. arXiv:2204.06125.

Ramesh A, Pavlov M, Goh G, Gray S, Voss C, Radford A, Chen M, Sutskever I (2021) *Zero-shot text-to-image generation*. arXiv:2102.12092

Rodriguez A, Granger R (2016) The grammar of mammalian brain capacity. Theoretical Computer Science C (TCS-C): 633:100-111. doi: 10.1016/j.tcs.2016.03.021

Rosenblatt F (1958) The perceptron: A probabilistic model for information storage and organization in the brain. *Psych Rev 65*:386-408

Rosenbloom P (2011). Rethinking cognitive architecture via graphical models. *Cog Systems Res., 12, 198-209.*

Rumelhart, D., & Zipser, D. (1986). Feature discovery by competitive learning. *Cognitive Science, 9*, 75-112.

Sampat S, Kumar A, Yang Y, Baral C (2021) CLEVR_HYP: A challenge dataset. Assoc Comp Ling, pp.3692-3709.

Serre, T. (2019). Deep learning: The good, the bad, and the ugly. *Ann Rev Vision Sci, 5*, 399-426.

Smolensky P, McCoy R, Fernandez R, Goldrick M, Gao J. (2022) *Neurocompositional computing: From the central paradox of cognition to a new generation of AI systems* arXiv:2205.01128v1

Thagard P, Shelley C (1997) Abductive reasoning. In Logic and scientific methods. Springer. pp.413-427.

Thrush, T., Jiang, R., Bartolo, M., Singh, A., Williams, A., Kiela, D., & Ross, C. (2022). *Winoground: Probing vision and language models for visio-linguistic compositionality*. arXiv:2204.03162v2

Vaswani A, Shazeer N, Parmar N, Uszkoreit J, Jones L, Gomez A, Kaiser L, Polosukhin I. (2017) Attention is all you need. NIPS 2017. arxiv.org/pdf/1706.03762.pdf

Werbos, P. (1974). *Beyond regression.* (Ph.D. diss). Harvard Univ

Widrow, B., & Hoff, M. (1960). *Adaptive switching circuits*. Stanford Univ Electronics Labs.

Zhou C, et al., (2023) A comprehensive survey on pretrained foundation models: A history from BERT to ChatGPT. arXiv:2302.09419v1

# A logical re-conception of neural networks: Hamiltonian bitwise part-whole architecture

## Supplemental Material

E.F.W.Bowen, R.Granger, A.Rodriguez
Dartmouth

**Abstract**

We describe a simple initial working system in which relations (such as part-whole) are directly represented via an architecture with operating and learning rules fundamentally distinct from artificial neural network methods. Arbitrary data are straightforwardly encoded as graphs whose edges correspond to codes from a small fixed primitive set of elemental pairwise relations, such that simple relational encoding is not an add-on, but occurs intrinsically within the most basic components of the system. A novel graph-Hamiltonian operator calculates energies among these encodings, with ground states denoting simultaneous satisfaction of all relation constraints among graph vertices. The method solely uses radically low-precision arithmetic; computational cost is correspondingly low, and scales linearly with the number of edges in the data. The resulting unconventional architecture can process standard ANN examples (two simple illustrative demonstrations are provided: MNIST, and credit applications) but also constructs representations of simple logical structures in these data (part-of; next-to) of a kind that is still highly elusive to typical ANNs. The resulting hierarchical representations enable abductive inferential steps generating relational position-based encodings rather than solely statistical representations. Notably, equivalent ANN operations are derived via a specific set of one-hot projections, identifying a special case of embedded vector encodings that may constitute a useful approach to current work in higher-level semantic representation. The very simple current state of the implemented system invites additional tools and improvements.

This supplemental material includes extended mathematical formulations, added tables, and additional figures illustrating results of running the working HNet system. All results in the paper and in these Supplemental results are generated by runnable code at:
https://github.com/DartmouthGrangerLab/hnet

## Appendix A: Initial part identification

For each given input, the bank of connected part components (from previous sections) produces a set of energies $\vec{y}$. Instead of providing these energies directly to a back-end classifier, we can provide them to a bank of tier 2 components. These meta components share a second, higher-tier graph $GT_{\mathbf{2}}$, in which each node (vertex) is a connected part component from the lower-tier graph $GT_{\mathbf{1}}$. Thus each edge in $GT_{\mathbf{2}}$ connects/associates two connected part components. As with connected part components, each vertex takes an activation, and each edge takes one of the sixteen states described previously. These edge states can be memorized or provided to learning algorithms.

It is important to note that an edge can only take on one of the four states if the nodes take on binary activations. Energies are non-negative real values, not binary numbers. For this purpose, a threshold or competitive nonlinearity is often applied to the energies, returning the values to binary and the edge states to one of our sixteen. However, certain learning algorithms can operate on the energies directly. These algorithms normalize the energies to the range 0 to 1, which then can be treated as probabilities. Probabilistic activations naturally yield probabilistic Hamiltonians. The probability that the state of edge (a,b) is (s1,s2) can be computed:

$$p( x_a = s_1 \text{ AND } x_b = s_2 ) = p(x_a = s_1)\, p(x_b = s_2)$$

## Appendix B: Image convolution/translation

In the previous sections, components were memorized — directly extracted from training data, then stored as-is. Beneficially, many existing and novel learning strategies can be applied at training time to improve accuracy, explainability, and efficiency of an HNet. For this first example, we will apply image convolution (a la CNNs, e.g., Fukushima & Miyake 1982, LeCun et al. 1989).

For image inputs, slight generalization is afforded by the application of a predetermined set of affine image transforms applied to all learned components, creating an expanded set that will recognize sub-images at a set of near locations. This enables a straightforward and inexpensive form of translation invariance. In the present example, we apply discrete rigid translation of each part at 2 pixels in the north, south, east, and west directions.

This is accomplished by formulating an appropriate permutation matrix $\boldsymbol{P}$ and applying it to each stored $\boldsymbol{H}$, to produce additional Hamiltonians ( $\boldsymbol{H'}$ ):

$$\boldsymbol{H'} = \boldsymbol{P}^{T} \boldsymbol{H} \boldsymbol{P}$$

Any arbitrary combination of translation or rotation may be used embedded in the permutation matrix using traditional image processing formulae.

The corresponding Hamiltonians $\boldsymbol{H'}$ are stored as new components. (Note that this step could instead be performed at recognition time, at greater time cost; the tradeoff has been to spend memory storage space to save time at recognition.)

## Appendix C: The sixteen atomic operators

As in **Figure 1a**, two inputs (images of the digit "4") that matched a given prototype may nonetheless have some different node and edge values. A given edge in a learned graph representation may thus have a value that is not one of the four simple binary pairs. For instance, if one of the 4's had a (0,1) edge at a location where another matching 4 had a (1,1) edge, then the value of the initial node at this edge has occurred once as a 0 and once as a 1. We may treat such instances in terms of logic statements: e.g., the value of the first node is indeterminate (either 0 or 1) whereas the second node must be a 1.

There are sixteen possible such combinations of binary pairs: see Supplemental Tables 1 and 2.

Four of these are the original binary pairs (**Table 3**); the remainder are the set of possible combinations of these four, as listed in
**Supplemental Table 1**.

If two adjacent graph nodes can each be either 1 or 0, such that the edge can be either (1,1) or (0,0) (but not (0,1) or (1,0)), then that edge can be logically described as an NXOR, which, as seen via the NXOR entry in
**Supplemental Table 1**, composes the binary combination of AND and NOR.

Similarly, any occurring combination of pairs that occur in data can contribute to the composition of a given edge as any of the sixteen possible combinations.

In practice, the state of a graph (the state of all of its edges) is often encoded as a vector $\vec{s}$ of length nedges. Each element corresponds to a particular edge, and its value is an integer {1 .. 16} identifying the state of the edge; the operator ID of the edge in
**Supplemental Table 1**. An ID of 0 indicates that the edge should be considered non-existent.

Using the same methods as in the previous section, the atomic Hamiltonians for all sixteen operators are readily derived (

$$\underline{+k = 0} \qquad\qquad \underline{+k = 1}$$

$$\hat{H}_{NOR} = \begin{bmatrix} 1 & -1/2 \\ -1/2 & 1 \end{bmatrix} \qquad \hat{H}_{OR} = \begin{bmatrix} -1 & 1/2 \\ 1/2 & -1 \end{bmatrix}$$

$$\hat{H}_{\neg X} = \begin{bmatrix} 1 & 0 \\ 0 & 0 \end{bmatrix} \qquad \hat{H}_{X} = \begin{bmatrix} -1 & 0 \\ 0 & 0 \end{bmatrix}$$

$$\hat{H}_{\neg Y} = \begin{bmatrix} 0 & 0 \\ 0 & 1 \end{bmatrix} \qquad \hat{H}_{Y} = \begin{bmatrix} 0 & 0 \\ 0 & -1 \end{bmatrix}$$

$$\hat{H}_{NAND} = \begin{bmatrix} 0 & 1/2 \\ 1/2 & 0 \end{bmatrix} \qquad \hat{H}_{AND} = \begin{bmatrix} 0 & -1/2 \\ -1/2 & 0 \end{bmatrix}$$

$$\hat{H}_{NXOR} = \begin{bmatrix} 1 & -1 \\ -1 & 1 \end{bmatrix} \qquad \hat{H}_{XOR} = \begin{bmatrix} -1 & 1 \\ 1 & -1 \end{bmatrix}$$

$$\hat{H}_{IMPL} = \begin{bmatrix} 1 & -1/2 \\ -1/2 & 0 \end{bmatrix} \qquad \hat{H}_{NIMPL} = \begin{bmatrix} -1 & 1/2 \\ 1/2 & 0 \end{bmatrix}$$

$$\hat{H}_{CONV} = \begin{bmatrix} 0 & -1/2 \\ -1/2 & 1 \end{bmatrix} \qquad \hat{H}_{NCONV} = \begin{bmatrix} 0 & 1/2 \\ 1/2 & -1 \end{bmatrix}$$

$$\hat{H}_{T} = \begin{bmatrix} 0 & 0 \\ 0 & 0 \end{bmatrix} \qquad \hat{H}_{F} = \begin{bmatrix} 0 & 0 \\ 0 & 0 \end{bmatrix}$$

**Supplemental Table 2**).

## Appendix D: Hamiltonian decomposition

For any vector of binary assignments of node values there exists a unique HNet Hamiltonian.

For a vector of node activations $\vec{x}$, and a graph where the relations among nodes are defined by the Hamiltonian ***H***, the energy ($E$) is given as:

$$E = x^T \ \hat{H} \ x + k$$

In the simplest case of a graph with two nodes and one edge, the Hamiltonian takes the form

$$\hat{H} = \begin{pmatrix} a & b \\ b & c \end{pmatrix}$$

where the possible values for ***H*** and *k* are given in Table D1 and depend on the particular logical relations between two nodes.

| a | b | c | k | edge | n1 | n2 |
|---|---|---|---|---|---|---|
| 1 | -1/2 | 1 | 0 | NOR | 0 | 0 |
| 0 | 1/2 | -1 | 1 | NC | 0 | 1 |
| -1 | 1/2 | 0 | 1 | NI | 1 | 0 |
| 0 | -1/2 | 0 | 1 | AND | 1 | 1 |

Table D1: Values {*a*, *b*, *c*, *k*} for the Hamiltonian energy equation, representing each of the logical relations (n1, n2).

For graphs containing multiple such edges, the corresponding Hamiltonians are composed by adding the unit Hamiltonians for the constituent edges, in their appropriate positions in the enlarged Hamiltonian, as shown in Section III (Equations 8 and 9) above. Edges can be defined by a relation between node i and node j where i < j, as in the following equation where r maps indices to a relation type:

$$r(i,j) : N \rtimes N \rightarrow \ \{NOR, NC, NI, AND\}$$

The values of the four variables {*a*, *b*, *c*, *k*} in Table D1 can thus be described in terms of four functions that can be used in the construction of a composite Hamiltonian.

$$a(r) : \{NOR, NC, NI, AND\} \rightarrow \{-1, 0, 1\}$$
$$b(r) : \{NOR, NC, NI, AND\} \rightarrow \{-1/2, 1/2\}$$
$$c(r) : \{NOR, NC, NI, AND\} \rightarrow \{-1, 0, 1\}$$
$$k(r) : \{NOR, NC, NI, AND\} \rightarrow \{0, 1\}$$

An HNet graph is then a set of tuples defining the *i* and *j* nodes that are connected as well as the relation type connecting them:

$$\text{edges} = \{(i, j, r) : i < j, r(i, j)\}$$

Given this set of edges, we construct a composite Hamiltonian ***H***$_{\text{composite}}$ as follows:

$$\hat{H}_{i,j} = \Sigma_{r\in\{(i,q,r)\in\text{edges}:i=j\}}\, a(r) \ + \Sigma_{r\in\{(i,j,r)\in\text{edges}:i\neq j\}}\, b(r)$$
$$+ \Sigma_{r\in\{(j,i,r)\in\text{edges}:i\neq j\}}\, b(r) \ + \Sigma_{r\in\{(p,j,r)\in\text{edges}:i=j\}}\, c(r)$$

Thus, for instance, in the four-edged graph from Figure 3, the H was composed by summing the four 2×2 Hamiltonians for the four edges, each sited in the appropriate spot in the composite Hamiltonian matrix ***H***$_{\text{composite}}$, as shown in Equations 8 and 9.

Table D2 shows the set of possible values for Hamiltonian matrix elements in the case of all possible consistent logical relation pairs:

| a1 | b1 | c1+a2 | b2 | c2 | e1 | e2 | n1 | n2 | n3 |
|---|---|---|---|---|---|---|---|---|---|
| 1 | -1/2 | 1+1=2 | -1/2 | 1 | NOR | NOR | 0 | 0 | 0 |
| 1 | -1/2 | 1+0=1 | 1/2 | -1 | NOR | NC | 0 | 0 | 1 |
| 0 | 1/2 | -1+1=-2 | 1/2 | 0 | NC | NI | 0 | 1 | 0 |
| 0 | 1/2 | -1+0=-1 | -1/2 | 0 | NC | AND | 0 | 1 | 1 |
| -1 | 1/2 | 0+1=1 | -1/2 | 1 | NI | NOR | 1 | 0 | 0 |
| -1 | 1/2 | 0+0=0 | 1/2 | -1 | NI | NC | 1 | 0 | 1 |
| 0 | -1/2 | 0-1=-1 | 1/2 | 0 | AND | NI | 1 | 1 | 0 |
| 0 | -1/2 | 0+0=0 | -1/2 | 0 | AND | AND | 1 | 1 | 1 |

Table D2. Possible values for Hamiltonian matrix elements for all consistent logical relation pairs.

Without any consistency constraints, a graph could theoretically be said to exist with edge relations that are not logically consistent with each other. For instance, if a hypothetical graph had two connected nodes as follows:

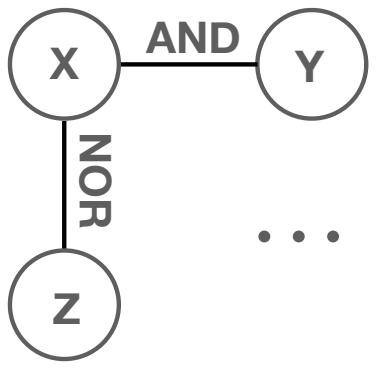

Then the AND edge would imply that *x* has a value of 1, whereas the NOR edge implies that *x* has a value of 0. These cannot simultaneously be true; the graph is logically inconsistent, i.e., no solution exists such that the constraints represented by these edges are satisfied. Ignoring all such cases of inconsistent graphs, the remaining logically consistent relations are those in Table D2.

For every consistent logical relation pair, there is a unique set of Hamiltonian matrix elements. There is thus a one-to-one mapping between the set of binary node assignments and their corresponding representation in the form of a Hamiltonian in an HNet.

## Appendix E: Equivalence of energy computation via $\hat{H}$ or via dimensional projections

The following extended pseudocode demonstrates the computation of the energy() function using Hamiltonians or using the projection of vectorized input into higher dimensions as described in section V in the main text ("Identification of equivalences").

```
G is the graph
n_cmp is the number of components (composite Hamiltonians)
compbankEdgeStates is G.n_edges × compbank.n_cmp (each element a 4-value enumeration)
data is a set of node values (0 or 1)
edgeData is G.n_edges × 1 (each element a 4-value enumeration)
```

```
function energies = Energy(G, n_cmp, compbankEdgeStates, data, do_h_mode)
    energies = zeros(n_cmp × 1)
    if do_h_mode
        for i = {1 .. n_cmp}
            [H,k] = GenerateCompositeH(G, compbankEdgeStates(:,cmp), i)
            energies(i) = data' * (H * data) + k
        energies = max(energies) - energies (convert from 0 = best to larger = better;
similarity)
    else
        edgeData = GetEdgeStates(data, G) (convert to edges)
        for i = {1 .. n_cmp}
            energies(i) = sum(compbankEdgeStates(:,i) == edgeData | compbankEdgeStates(:,i) ==
NULL)
        energies = energies – min(energies)
```

```
function [H,k] = GenerateCompositeH(G, edgeStates, cmp)
    n_present_edges = sum(edgeStates ≠ NULL)
    rows = zeros(n_present_edges * 3 × 1);
    cols = zeros(n_present_edges * 3 × 1);
    vals = zeros(n_present_edges * 3 × 1);

    op = [1 -1  1  0;... (NOR)
          0  1 -1  1;... (NCONV)
         -1  1  0  1;... (NIMPL)
          0 -1  0  1]'   (AND)
    k = 0
    count = 0
    for r = {1 .. 4} (for each unit H type)
        mask = (edgeStates == r)
        n_new = sum(mask)

        edgeNodes = G.edgeEndnodeIdx(mask,:) (n_edges × 2; src and dst node)

        rows(count+(1:n_new)) = edgeNodes(:,1)
        cols(count+(1:n_new)) = edgeNodes(:,1)
        vals(count+(1:n_new)) = op(1,r) (implicit expansion)
        count = count + n_new

        rows(count+(1:n_new)) = edgeNodes(:,1)
        cols(count+(1:n_new)) = edgeNodes(:,2)
        vals(count+(1:n_new)) = op(2,r) (implicit expansion)
        count = count + n_new

        rows(count+(1:n_new)) = edgeNodes(:,2)
        cols(count+(1:n_new)) = edgeNodes(:,2)
        vals(count+(1:n_new)) = op(3,r) (implicit expansion)
        count = count + n_new
```

```
        k = k + op(4,r) * n_new
    H = sparse(rows, cols, vals, G.n_nodes, G.n_nodes)
```

```
function edgeStates = GetEdgeStates(data, G)
    temp = data(G.edgeEndnodeIdx(:,1)) .* 2 + data(G.edgeEndnodeIdx(:,2))
    edgeStates = zeros(G.n_edges × 1)
    edgeStates(temp == 0) = 1 (NOR)
    edgeStates(temp == 1) = 2 (NCONV)
    edgeStates(temp == 2) = 3 (NIMPL)
    edgeStates(temp == 3) = 4 (AND)
    edgeStates = FilterEdgeType(edgeStates, G.edgeTypeFilter)
```

## SUPPLEMENTAL TABLES

| Operator | s<br>State ID | (1,1)<br>AND | (1,0)<br>NIMPL | (0,1)<br>NCONV | (0,0)<br>NOR |
|---|---|---|---|---|---|
| NULL | - | - | - | - | - |
| false | 0 | 0 | 0 | 0 | 0 |
| NOR | 1 | 0 | 0 | 0 | 1 |
| NCONV | 2 | 0 | 0 | 1 | 0 |
| ¬X | 3 | 0 | 0 | 1 | 1 |
| NIMPL | 4 | 0 | 1 | 0 | 0 |
| ¬Y | 5 | 0 | 1 | 0 | 1 |
| XOR | 6 | 0 | 1 | 1 | 0 |
| NAND | 7 | 0 | 1 | 1 | 1 |
| AND | 8 | 1 | 0 | 0 | 0 |
| NXOR | 9 | 1 | 0 | 0 | 1 |
| Y | 10 | 1 | 0 | 1 | 0 |
| IMPL | 11 | 1 | 0 | 1 | 1 |
| X | 12 | 1 | 1 | 0 | 0 |
| CONV | 13 | 1 | 1 | 0 | 1 |
| OR | 14 | 1 | 1 | 1 | 0 |
| true | 15 | 1 | 1 | 1 | 1 |

**Supplemental Table 1.** The sixteen logical operator combinations, in terms of the four fundamental operators based on {(1,1), (1,0), (0,1), (0,0)}, or {(true,true), (true,false), (false,true), (false,false)}. The utility of these operators is twofold: i) their combinatory semantics in Hamiltonian operations; ii) their manipulability by bitwise-AND and bitwise-OR operations (see text).

| $+k = 0$ | $+k = 1$ |
|---|---|
| $\hat{H}_{NOR} = \begin{bmatrix} 1 & -1/2 \\ -1/2 & 1 \end{bmatrix}$ | $\hat{H}_{OR} = \begin{bmatrix} -1 & 1/2 \\ 1/2 & -1 \end{bmatrix}$ |
| $\hat{H}_{\neg X} = \begin{bmatrix} 1 & 0 \\ 0 & 0 \end{bmatrix}$ | $\hat{H}_{X} = \begin{bmatrix} -1 & 0 \\ 0 & 0 \end{bmatrix}$ |
| $\hat{H}_{\neg Y} = \begin{bmatrix} 0 & 0 \\ 0 & 1 \end{bmatrix}$ | $\hat{H}_{Y} = \begin{bmatrix} 0 & 0 \\ 0 & -1 \end{bmatrix}$ |
| $\hat{H}_{NAND} = \begin{bmatrix} 0 & 1/2 \\ 1/2 & 0 \end{bmatrix}$ | $\hat{H}_{AND} = \begin{bmatrix} 0 & -1/2 \\ -1/2 & 0 \end{bmatrix}$ |
| $\hat{H}_{NXOR} = \begin{bmatrix} 1 & -1 \\ -1 & 1 \end{bmatrix}$ | $\hat{H}_{XOR} = \begin{bmatrix} -1 & 1 \\ 1 & -1 \end{bmatrix}$ |
| $\hat{H}_{IMPL} = \begin{bmatrix} 1 & -1/2 \\ -1/2 & 0 \end{bmatrix}$ | $\hat{H}_{NIMPL} = \begin{bmatrix} -1 & 1/2 \\ 1/2 & 0 \end{bmatrix}$ |
| $\hat{H}_{CONV} = \begin{bmatrix} 0 & -1/2 \\ -1/2 & 1 \end{bmatrix}$ | $\hat{H}_{NCONV} = \begin{bmatrix} 0 & 1/2 \\ 1/2 & -1 \end{bmatrix}$ |
| $\hat{H}_{T} = \begin{bmatrix} 0 & 0 \\ 0 & 0 \end{bmatrix}$ | $\hat{H}_{F} = \begin{bmatrix} 0 & 0 \\ 0 & 0 \end{bmatrix}$ |

**Supplemental Table 2.** Hamiltonians for all 16 operators. *+k* values indicate value (1 or 0) added to specified Hamiltonian for energy calculation (see text).

| feature | interpretation | % of time feature appears in datapoints of class: "good" | "bad" |
|---|---|---|---|
| a1_A11 | salary for at least 1 year was ≤ DM | 8.67% | 23.33% |
| a1_A12 | salary for at least 1 year was < 200 DM | 11.00% | 16.33% |
| a1_A13 | salary for at least 1 year was > 200 DM | 3.67% | 2.67% |
| a1_A14 | no checking acct | 26.67% | 7.67% |
| a3_A30 | no credits taken / all credits paid back duly | 0.67% | 4.67% |
| a3_A31 | all credits at this bank paid back duly | 1.00% | 4.67% |
| a3_A32 | existing credits paid back duly till now | 25.67% | 28.00% |
| a3_A33 | delay in paying off in the past | 4.67% | 5.33% |
| a3_A34 | critical account / other credits existing (not at this bank) | 18.00% | 7.33% |
| a4_A40 | purpose = car (new) | 9.67% | 15.00% |
| a4_A41 | purpose = car (used) | 8.33% | 2.33% |
| a4_A42 | purpose = furniture/equipment | 8.67% | 10.00% |
| a4_A43 | purpose = radio/television | 16.33% | 12.67% |
| a4_A44 | purpose = domestic appliances | 0.33% | 0.33% |
| a4_A45 | purpose = repairs | 1.00% | 2.00% |
| a4_A46 | purpose = education | 1.33% | 2.00% |
| a4_A48 | purpose = retraining | 0.33% | 0 |
| a4_A49 | purpose = business | 4.00% | 5.00% |
| a4_A410 | purpose = other | 0 | 0.67% |
| a6_A61 | savings and bonds = < 100 dm | 26.67% | 35.00% |
| a6_A62 | savings and bonds = 100 to 500 dm | 5.67% | 7.00% |
| a6_A63 | savings and bonds = 500 to 1000 dm | 3.67% | 2.00% |
| a6_A64 | savings and bonds = over 1000 dm | 4.67% | 0.67% |
| a6_A65 | savings and bonds = unknown/none | 9.33% | 5.33% |
| a7_A71 | unemployed | 2.67% | 4.00% |
| a7_A72 | present job held for < 1 year | 6.67% | 10.33% |
| a7_A73 | present job held for 1-4 years | 17.33% | 16.67% |
| a7_A74 | present job held for 4-7 years | 8.67% | 8.00% |
| a7_A75 | present job held for > 7 years | 14.67% | 11.00% |
| a9_A91 | male, divorced/separated | 2.33% | 4.00% |
| a9_A92 | female, divorced/separated/married | 12.33% | 19.67% |
| a9_A93 | male, single | 31.33% | 24.00% |
| a9_A94 | male, married/widowed | 4.00% | 2.33% |
| a10_A101 | other debtors/guarantors = none | 45.67% | 44.67% |
| a10_A102 | other debtors/guarantors = co-applicant | 1.33% | 3.67% |
| a10_A103 | other debtors/guarantors = guarantor | 3.00% | 1.67% |
| a12_A121 | property = real estate | 17.33% | 10.33% |
| a12_A122 | property = if not A121: building society savings agreement / life | 11.00% | 11.00% |
| a12_A123 | insurance | 17.00% | 17.00% |
| a12_A124 | property = if not A121/A122: car or other, not in attribute 6 | 4.67% | 11.67% |
| a14_A141 | property = unknown/none | 5.33% | 10.00% |
| a14_A142 | other installment plans = bank | 3.00% | 1.33% |
| a14_A143 | other installment plans = stores | 41.67% | 38.67% |
| a15_A151 | other installment plans = none | 8.00% | 11.67% |
| a15_A152 | housing = rent | 38.33% | 32.00% |
| a15_A153 | housing = own | 3.67% | 6.33% |
| a17_A171 | housing = for free | 1.33% | 1.33% |
| a17_A172 | job = unemployed/unskilled, non-resident | 10.33% | 7.67% |
| a17_A173 | job = unskilled, resident | 30.67% | 33.67% |
| a17_A174 | job = skilled employee / official | 7.67% | 7.33% |
| a2_1 | job = management / self-employed / highly qualified employee / | 30.67% | 22.67% |
| a2_2 | officer | 13.67% | 13.00% |
| a2_3 | duration, bucket 1 | 4.33% | 7.33% |
| a2_4 | duration, bucket 2 | 0.67% | 5.33% |
| a2_5 | duration, bucket 3 | 0.67% | 1.67% |
| a5_1 | duration, bucket 4 | 42.00% | 37.00% |
| a5_2 | duration, bucket 5 | 6.33% | 8.00% |
| a5_3 | credit score, bucket 1 | 1.33% | 3.00% |
| a5_4 | credit score, bucket 2 | 0.33% | 1.67% |
| a5_5 | credit score, bucket 3 | 0 | 0.33% |
| a8_1 | credit score, bucket 4 | 6.00% | 6.00% |
| a8_2 | credit score, bucket 5 | 15.33% | 10.33% |
| a8_3 | percent = "1" | 6.00% | 7.33% |
| a8_4 | percent = "2" | 22.67% | 26.33% |
| a11_1 | percent = "3" | 5.00% | 6.67% |
| a11_2 | percent = "4" | 17.005 | 16.00% |
| a11_3 | present residence since = "1" | 7.33% | 9.00% |
| a11_4 | present residence since = "2" | 20.67% | 18.33% |
| a13_1 | present residence since = "3" | 23.33% | 32.33% |
| a13_2 | present residence since = "4" | 17.00% | 9.67% |
| a13_3 | age, bucket 1 | 5.67% | 5.00% |
| a13_4 | age, bucket 2 | 3.67% | 3.00% |
| a13_5 | age, bucket 3 | 0.33% | 0 |
| a16_1 | age, bucket 4 | 29.00% | 32.00% |
| a16_2 | age, bucket 5 | 19.00% | 16.67% |
| a16_3 | number of credits, bucket 1 | 1.67% | 0.67% |
| a16_4 | number of credits, bucket 2 | 0 | 0 |
| a16_5 | number of credits, bucket 3 | 0.33% | 0.67% |
| a18 | number of credits, bucket 4 | 8.33% | 8.33% |
| a19 | number of credits, bucket 5 | 22.00% | 15.67% |
| a20 | has a dependent | 47.67% | 49.00% |
| | has phone | | |
| | is foreign | | |

**Supplemental Table 3.** Description of each (binarized) feature in the consumer credit application dataset (see main paper). For each feature, we calculated the percentage of the time that it appears (is a 1 not a 0) in training datapoints labeled "good," and separately in datapoints labeled "bad."

input: $\vec{x}_{\text{tst}}$, a single datapoint withheld from training load: $\widehat{H}_{\text{tier1}}$, $\vec{k}_{\text{tier1}}$, ($\widehat{H}_{\text{tier2}}$, $\vec{k}_{\text{tier2}}$), *model*

$\vec{y} \leftarrow$ ENCODE($\widehat{H}_{\text{tier1}}$, $\vec{k}_{\text{tier1}}$, $\vec{x}_{\text{tst}}$)

$\vec{y} \leftarrow$ ENCODE($\widehat{H}_{\text{tier2}}$, $\vec{k}_{\text{tier2}}$, $\vec{y}$) (for multi-tier networks)

$\vec{y} \leftarrow$ PREDICT(*model*, $\vec{y}$)

**Supplemental Table 4.** Testing algorithm. (See text).

$\vec{y}_{\text{similarity}} \leftarrow \text{ENCODE}_{\text{similarity}}(\boldsymbol{R}, \vec{x}_{\text{tst}}, G)$ is:

$\vec{s} \in$ ℕnedges

for $i \in \{1 .. n_{\text{edges}}\}$ (convert activations to edge states)

$s_i \leftarrow$ LOOKUP(($x_{\text{tst}}$ [$G$.edge_$i$_node_1], $x_{\text{tst}}$ [$G$.edge_$i$_node_2])) (see Table 3)

for $i \in \{1 .. n_{\text{cmp}}\}$

$y_i \leftarrow$ sum($R$[:, $i$] == $\vec{s}$ | $R$[:, $i$] == NULL)

**Supplemental Table 5.** Alternate encoding function.

## SUPPLEMENTAL FIGURES

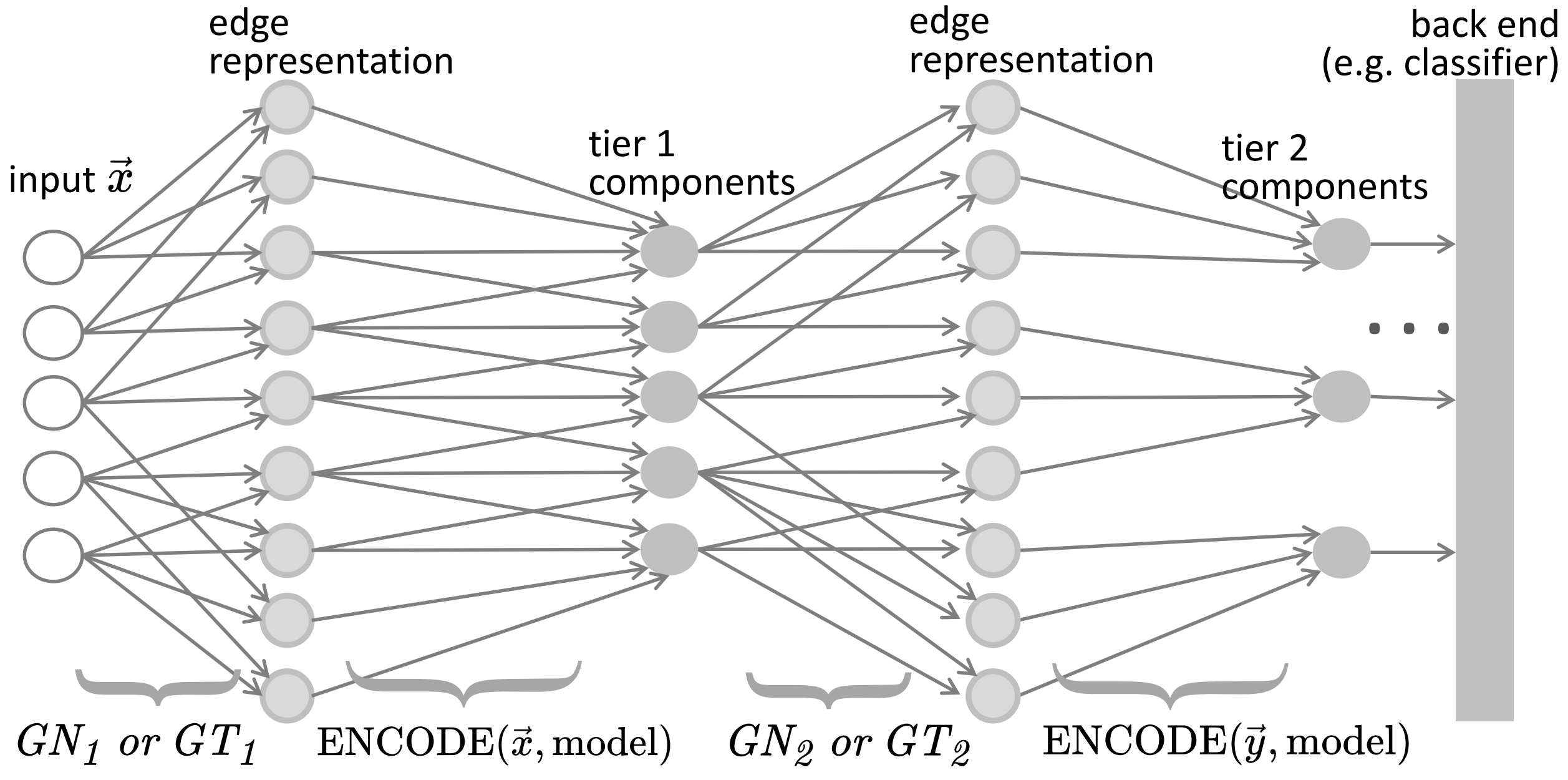

**Supplemental Figure 1.** (See Figure 9 in main text). Abstract structure of the two simple HNets presented in the main text. The architecture applies both to topographic and nontopographic input data, as in the MNIST and Credit card examples, respectively.

A topographic input (e.g., MNIST or other visual or auditory data) contains adjacency structure information as encoded in the form of a GT1 directed graph (see, e.g., Figures 1 and 3 in the main text). Each edge in the input graph takes one of four states, given the values of the pixels an edge connects. Learned components (see text) are compared with the values of edges in the input graph, producing a similarity score; the resultant similarity scores can be used for classification, or presented as input to a back end (possibly very simple) classifier.

A nontopographic input (e.g., credit card application or other abstract data) contains value-attribute pairs, as in normal machine learning or neural network applications; these are formatted into the GN1 graph format (see Figures 6 and 7 in the main text, and Supplemental figures 10-12 below).

For both topographic and nontopographic data, once components have been identified in the input, these can be further processed as inputs and higher level structures created (GN2 or GT2 graph formats); the resulting heterarchies can continue to higher tiers.

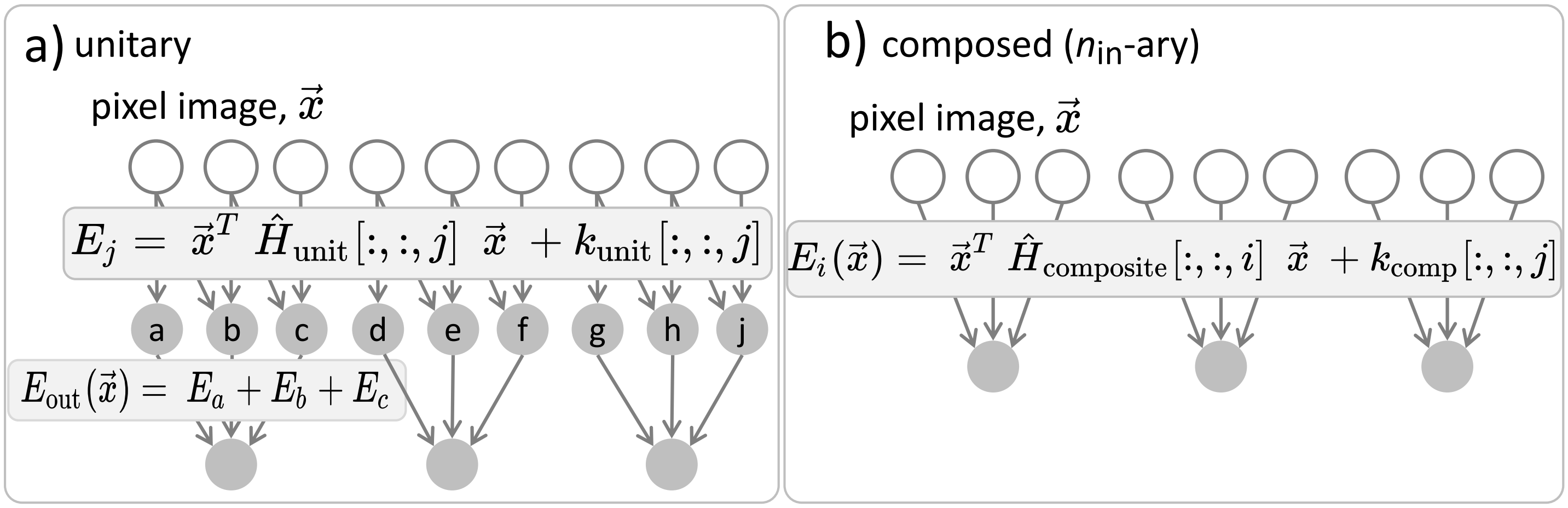

**Supplemental Figure 2.** Two image-processing networks that produce identical results. **a)** An HNet comprised exclusively of unit (single-edge) Hamiltonians and summation components. **b)** An HNet comprised exclusively of composite Hamiltonians. All components perform the same operation.

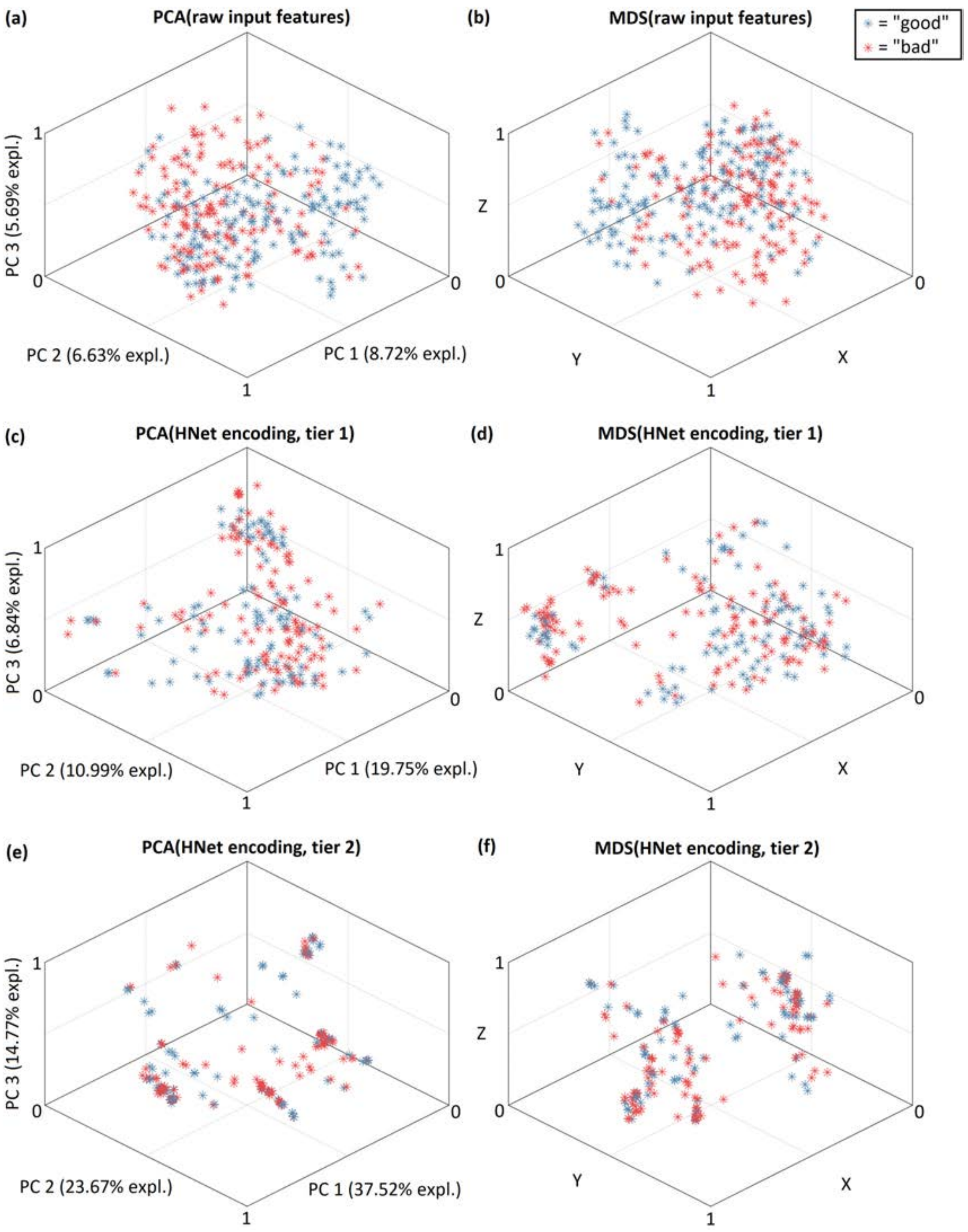

**Supplemental Figure 3.** Testing half of the consumer credit application dataset. Each datapoint is embedded in a common three-dimensional space (first three principal components **(left)** and three-dimensional multidimensional scaling **(right)**). The dataset provides correct labels, "good" (assumed to mean "accept") in blue and "bad" (assumed to mean "reject") in red. **a, b)** Embedding derived from the presence or absence of the binary input features in

**Supplemental Table 3**. **c, d)** Embedding derived from the energies of the tier 1 Hamiltonians. **e, f)** Embedding derived from the energies of the tier 2 Hamiltonians.

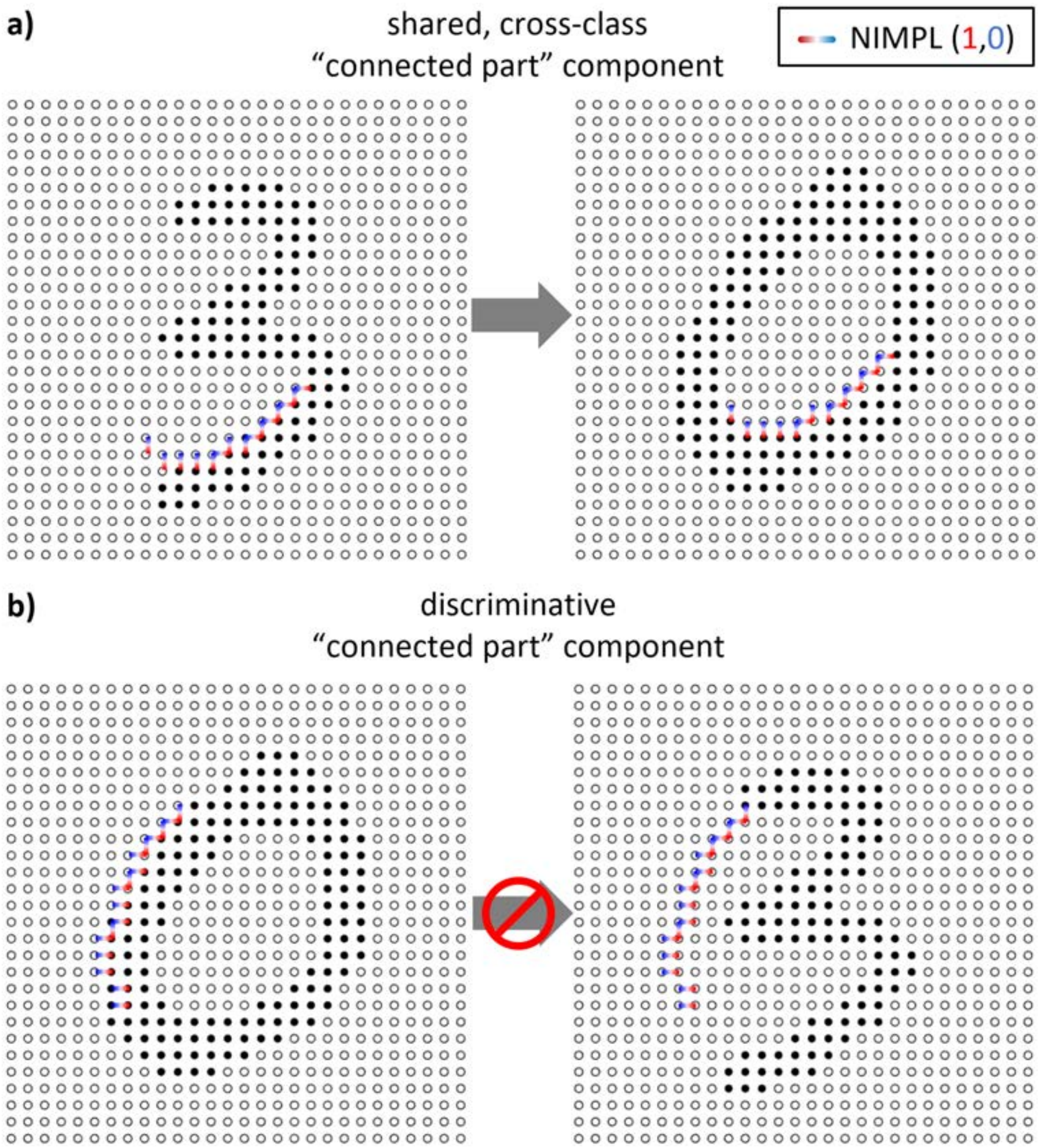

**Supplemental Figure 4. a)** Some digit classes share components, like the one presented on this "3" and this "0," in common. **b)** Some digit classes can be discriminated using a small number of components, like this component that matches the "0" but not the "3."

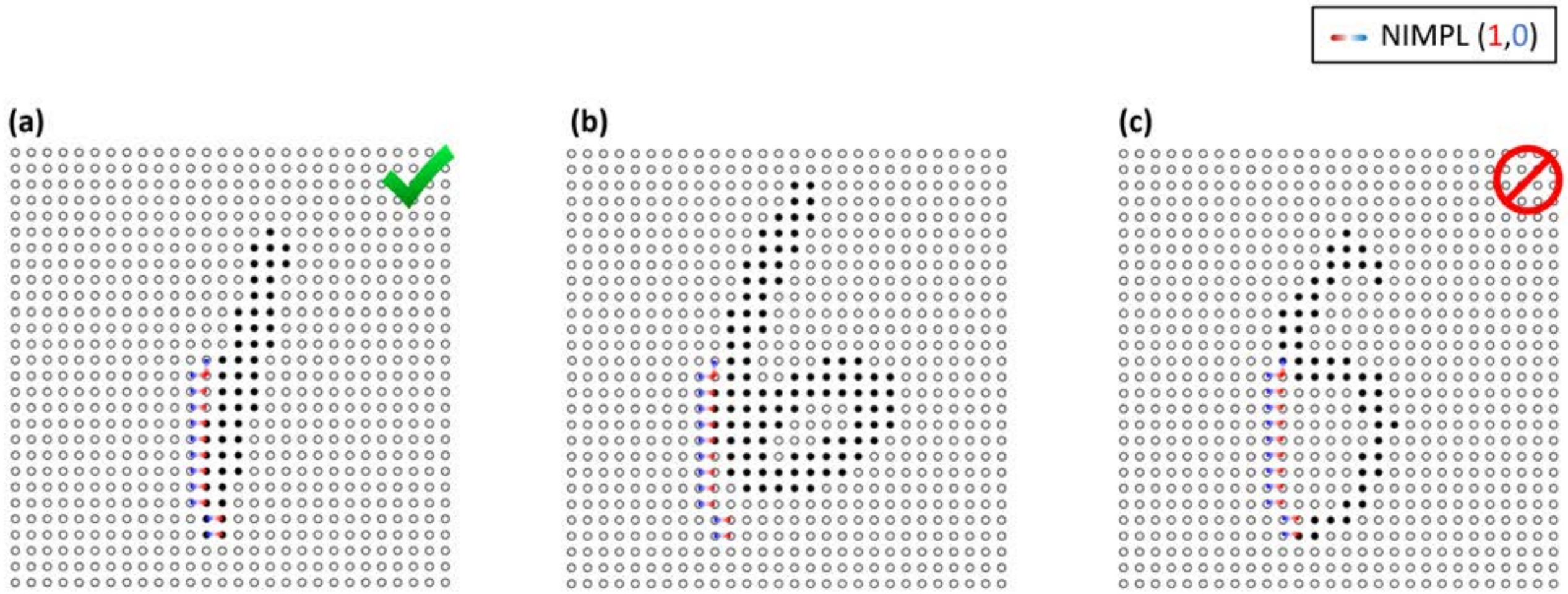

**Supplemental Figure 5.** Example component-image matches, MNIST dataset. In reality, features may produce a variety of energies. **a)** The component matches this “1” well. **b)** The component matches this “6” somewhat. **c)** The component matches this “5” very poorly.

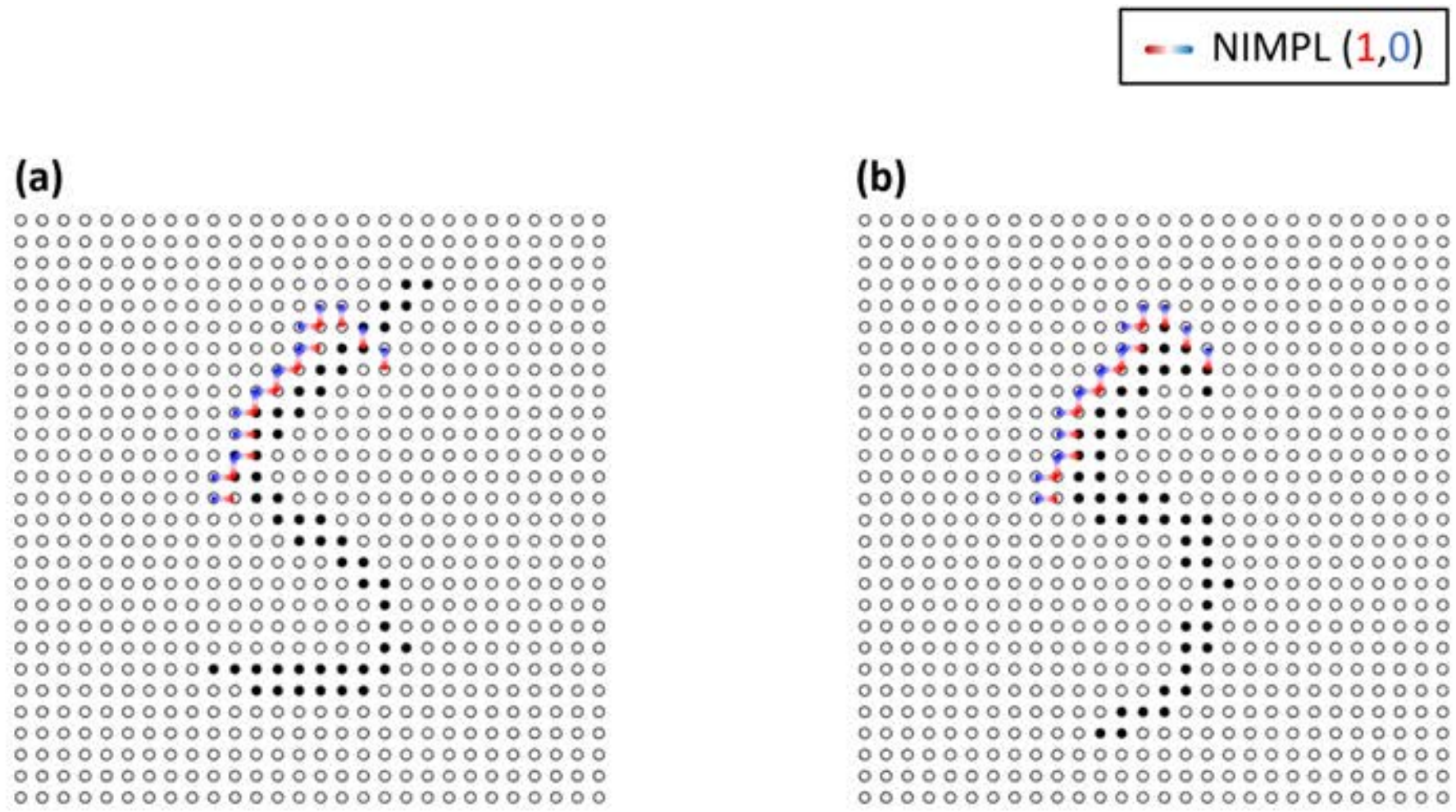

**Supplemental Figure 6.** Example component-image matches, MNIST dataset. The presented component matches **a)** some hand-written instances of a "5" but not others **b)**.

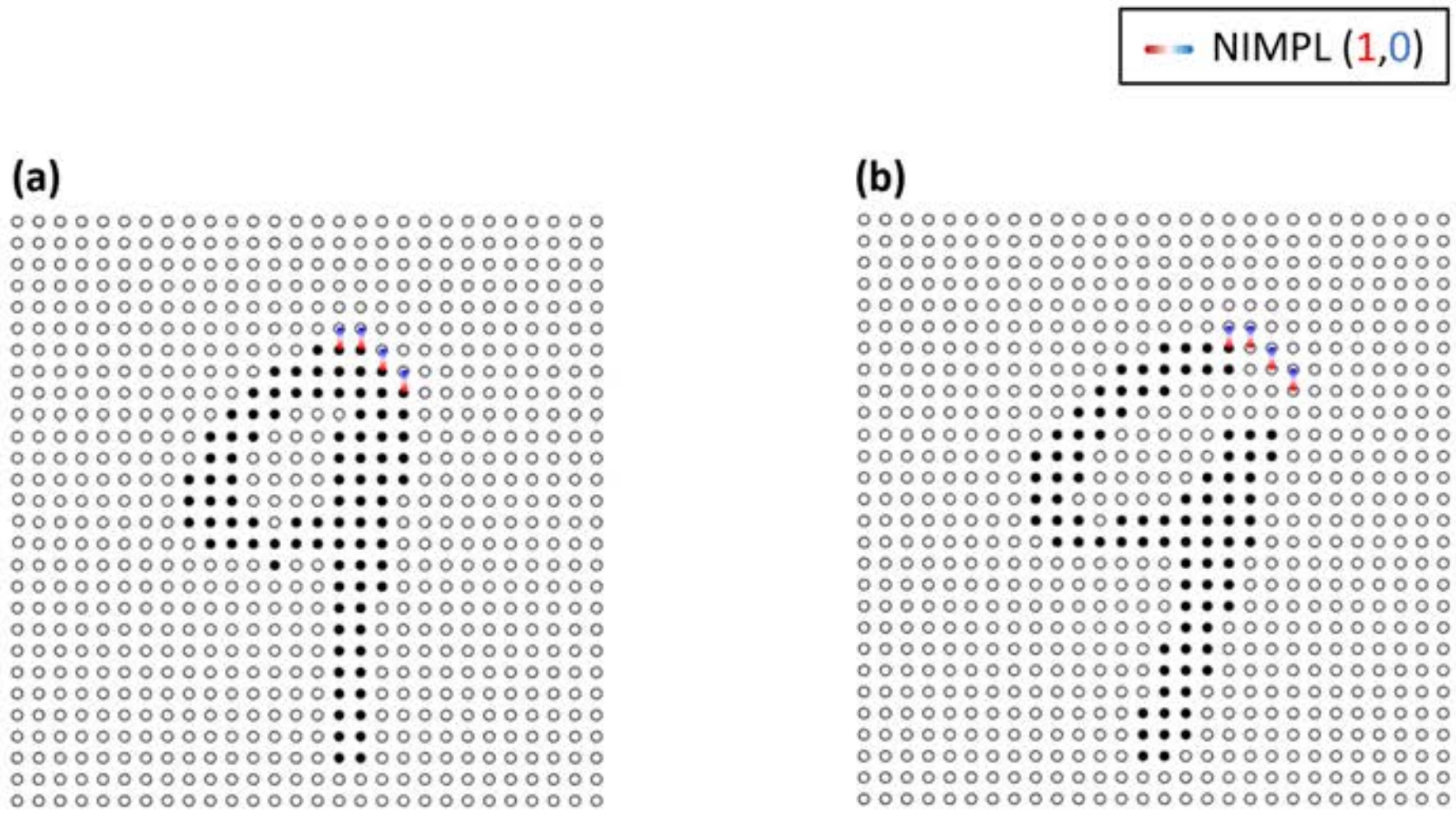

**Supplemental Figure 7.** Example component-image matches, MNIST dataset. Overly-short components are prone to mistakes. For example, the presented 4-edge part may match one “9” **a)** and not another **b)**.

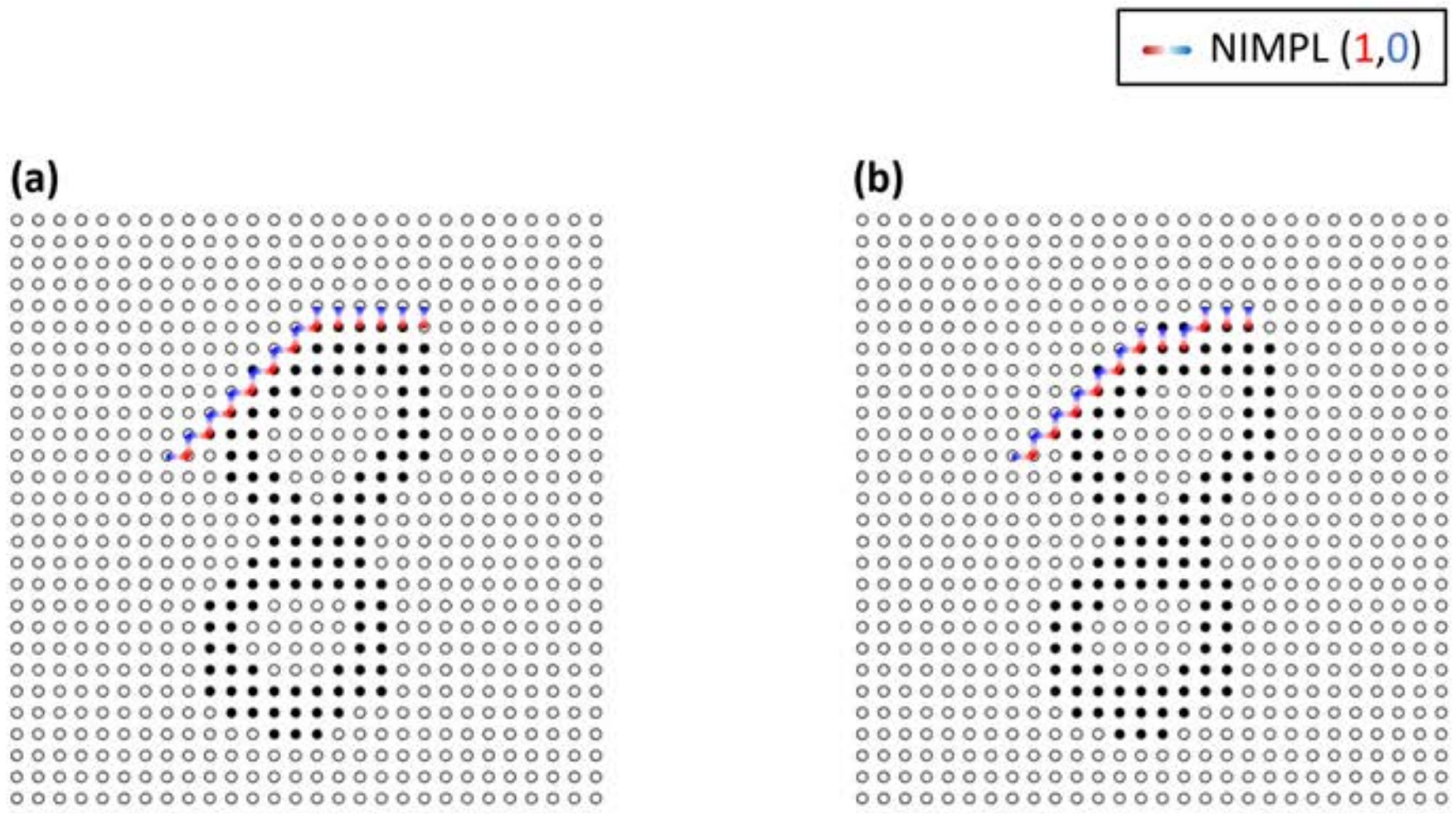

**Supplemental Figure 8.** Example component-image matches, MNIST dataset. Some components may match similar set of images (their energies may correlate). For example, we present two very similar parts, **a)** and **b)**, on the same image.

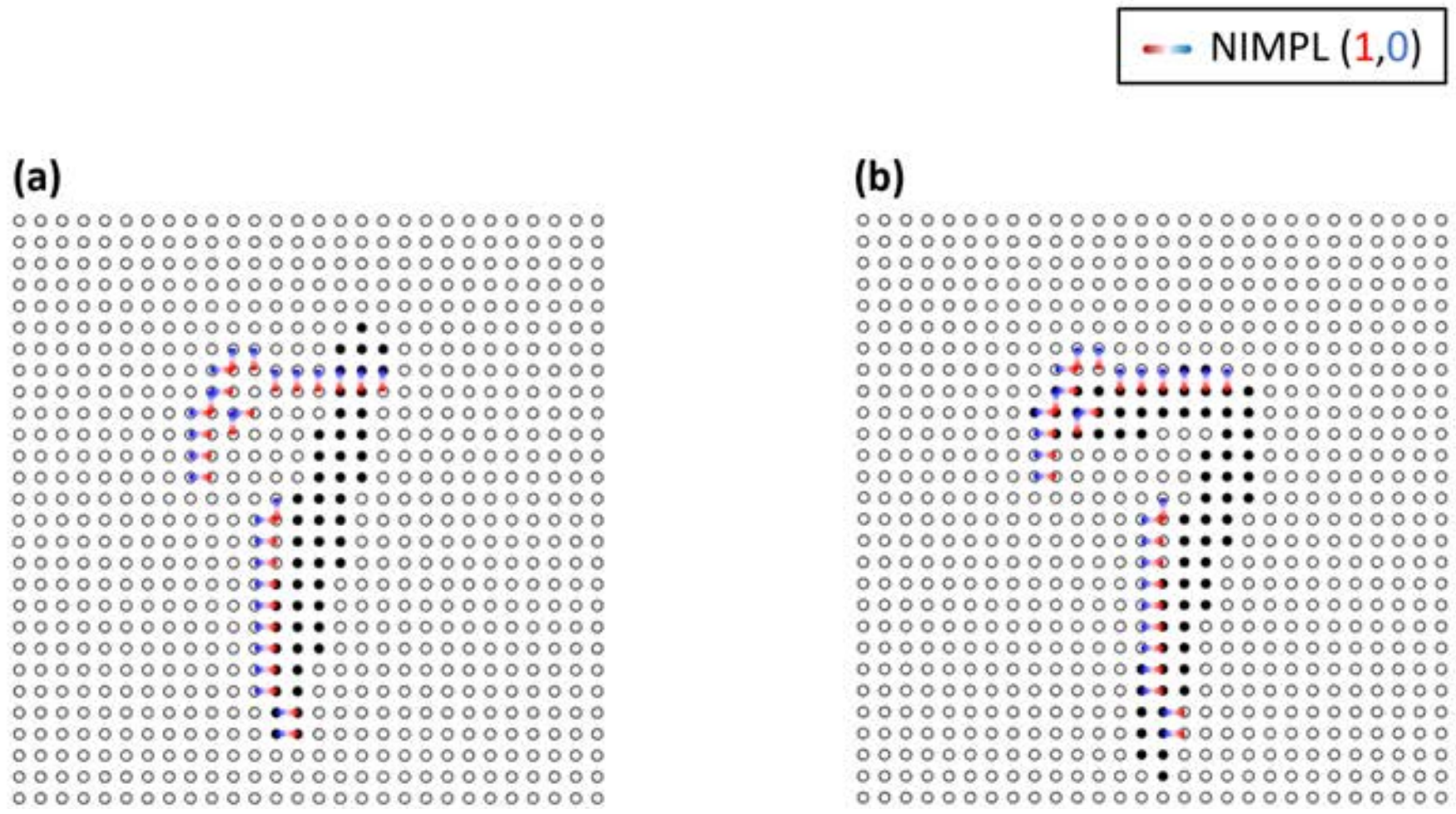

**Supplemental Figure 9.** Example component-image matches, MNIST dataset. Here, we present **a)** a "1" and **b)** a "7." We match the same two connected part components to each. The lower connected part component is far less discriminative than the one on top.

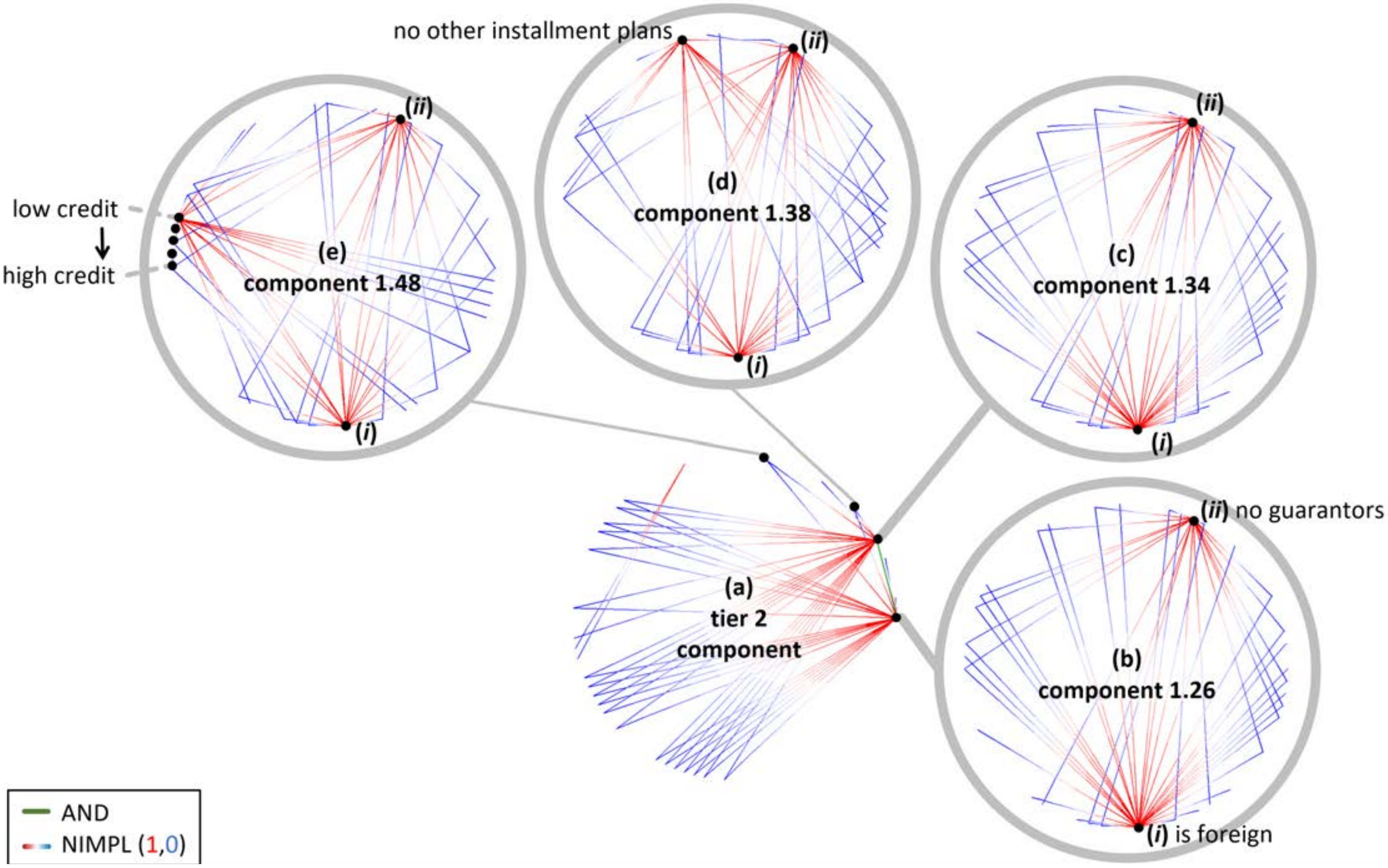

**Supplemental Figure 10.** Interpretation of **a)** a tier 2 component. Each graph is represented as a circle of nodes; only those explained are drawn. Lines between nodes are draw if present in the component's definition. AND edges are drawn in green, NIMPL edges are drawn with the active (1) node in red, and the inactive (0) node in blue. The thickness of gray lines illustrates the approximate number of tier 2 edges in which the tier 1 component is a participant. For tier 1 components, nodes are binarized credit application features. For tier 2 components, nodes are tier 1 components. This tier 2 component is dominated by the requirement that two tier 1 components b,c) both produce good energy. This is indicated by the AND relation between them, and by the number of NIMPL relations between b,c) and other tier-1 components. Components **b)** and **c)** are very similar — they produce good energy when a credit applicant is both foreign and without guarantors. By contrast, this tier 2 component prefers that tier 1 components d) and e) are not present (0). Component **d)** also prefers foreigners without guarantors, but specifies those who have no other installment plans. Component **e)** prefers foreigners without guarantors, but specifies those with low credit.

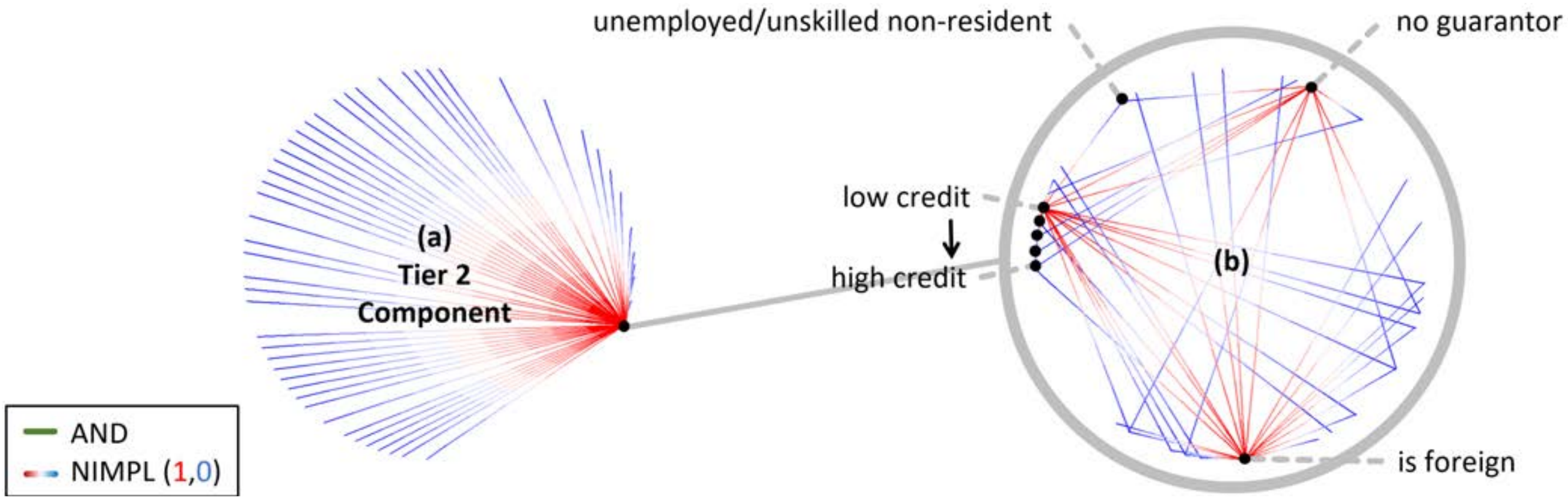

**Supplemental Figure 11.** Interpretation of **a)** a second tier 2 component, formatted as in **Supplemental Figure 10**. The thickness of gray lines illustrates the approximate number of tier 2 edges in which the tier 1 component is a participant. This component focuses on one particular tier 1 component. **b)** The tier 1 component seems to represent foreigners with no guarantor and low credit, but who are not unemployed/unskilled.

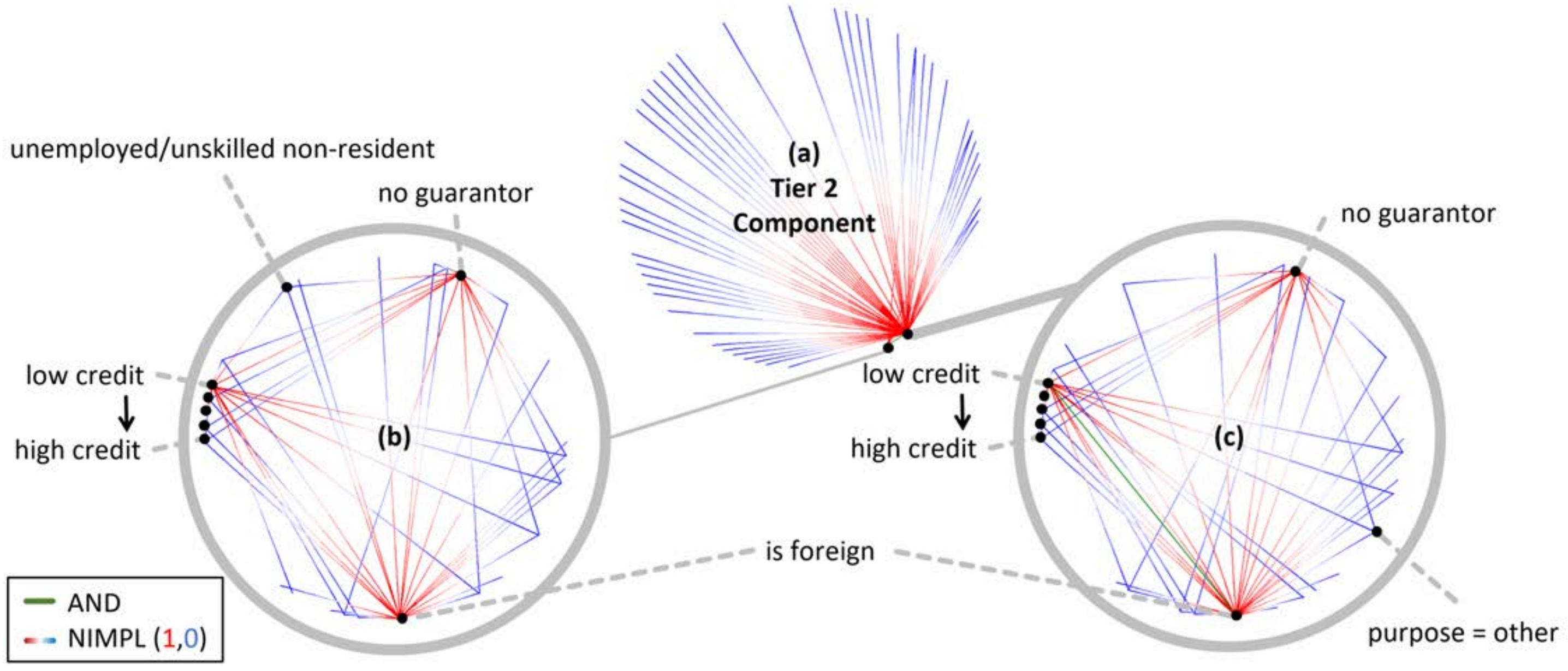

**Supplemental Figure 12.** Interpretation of **a)** a third tier 2 component, formatted as in **Supplemental Figure 10**. The thickness of gray lines illustrates the approximate number of tier 2 edges in which the tier 1 component is a participant. This component produces the best energy in response to datapoints for which both tier 1 components b) and c) are true, though it prefers c). **b)** This tier 1 component seems to represent foreigners with no guarantor and low credit, but who are not unemployed/unskilled. **c)** This tier 1 component is slightly less responsive to employment status, but more strongly associates foreign status with low credit.

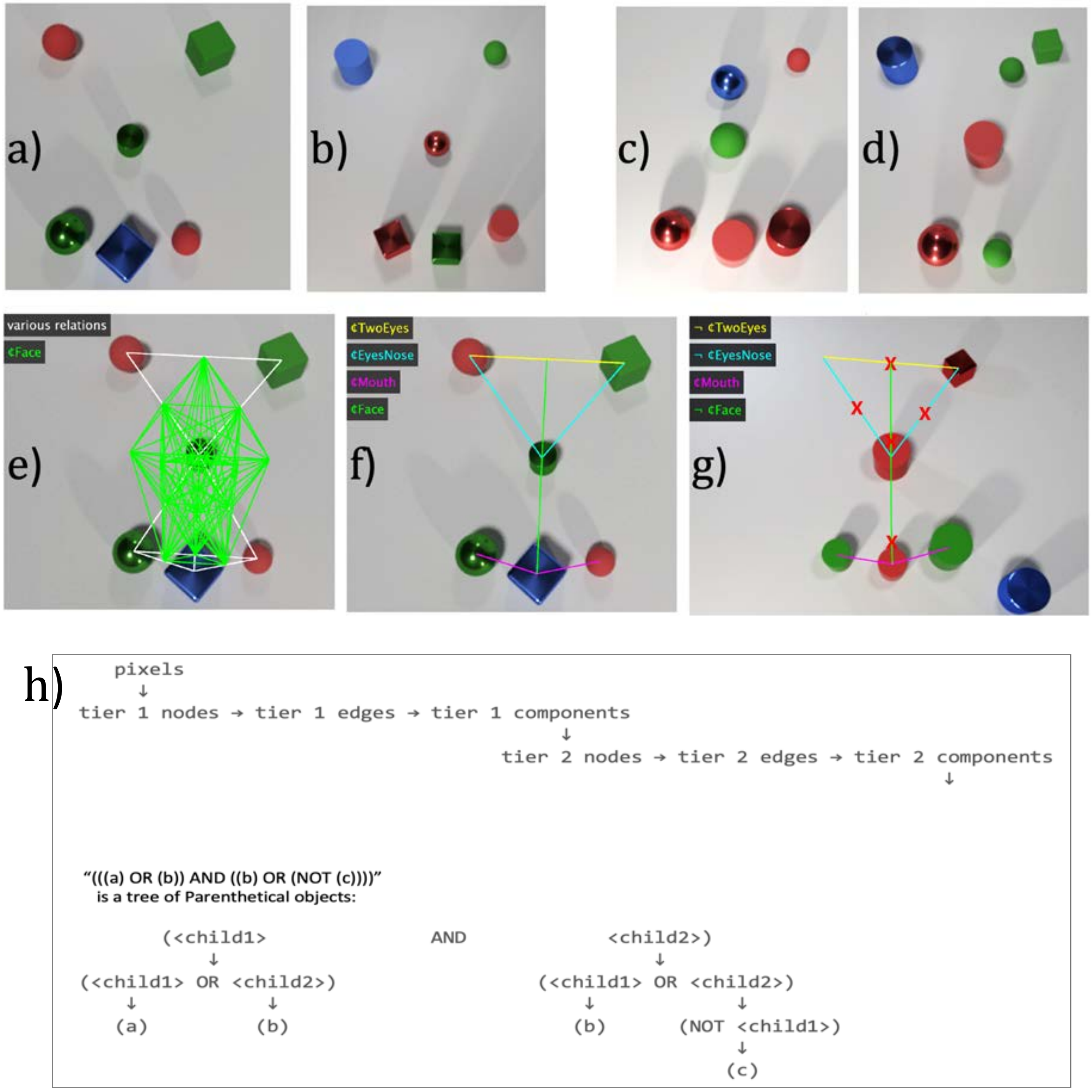

**Supplemental Figure 13.** Sample images generated in the CLEVR system (Sampat et al., 2021). Configurations are presented (a few examples in a – d) and acquired into initial large groups of instances, e.g., (e), which are examined for shared regularities (h) and trained to match e.g., (f) or mismatch (g).